%% file: AAAI26_cameraready.tex
\documentclass[letterpaper]{article} 
\usepackage{aaai2026}  
\usepackage{times}  
\usepackage{helvet}  
\usepackage{courier}  
\usepackage[hyphens]{url}  
\usepackage{graphicx} 
\usepackage{natbib}  
\usepackage{caption} 
\usepackage{algorithm}
\usepackage{algorithmic}
\usepackage[inline]{enumitem}
\usepackage{amsmath}
\usepackage{amsthm}
\usepackage{thm-restate}
\usepackage{amssymb}
\usepackage{thmtools}
\usepackage{mathtools}
\usepackage[capitalise]{cleveref}
\usepackage[all]{foreign}

\usepackage{placeins}

\newtheorem{definition}{Definition}

\usepackage{newfloat}
\usepackage{listings}
\DeclareCaptionStyle{ruled}{labelfont=normalfont,labelsep=colon,strut=off} 
\floatstyle{ruled}
\newfloat{listing}{tb}{lst}{}
\floatname{listing}{Listing}
\usepackage{etoolbox}
\usepackage{xspace}
\NewDocumentCommand\DeclareSymbol{O{} o m}{%
  \IfValueTF{#2}{%
    \def#2{\ensuremath{#1{#3}}\xspace}%
  }{%
    \csdef{#3}{\ensuremath{#1{#3}}\xspace}%
  }%
}

\DeclareSymbol[\mathcal]{V}
\DeclareSymbol[\mathcal]{C}
\DeclareSymbol[\mathcal]{F}
\DeclareSymbol[\mathcal]{R}
\DeclareSymbol[\mathbb]{D}

\NewDocumentCommand\choice{}{\mathrel{\vert}}

\usepackage{multirow}
\usepackage{booktabs}
\usepackage[flushleft]{threeparttable}

\usepackage{xcolor}

\usepackage{amsmath}
\DeclareMathOperator{\Tet}{Tet}
\DeclareMathOperator{\Cube}{Cube}
\DeclareMathOperator{\Dodec}{Dodec}

\DeclareMathOperator{\LeftOf}{LeftOf}
\DeclareMathOperator{\BackOf}{BackOf}
\DeclareMathOperator{\FrontOf}{FrontOf}
\DeclareMathOperator{\Medium}{Medium}
\DeclareMathOperator{\Small}{Small}
\DeclareMathOperator{\Larger}{Larger}
\DeclareMathOperator{\Smaller}{Smaller}

\DeclareMathOperator{\Country}{Country}
\DeclareMathOperator{\InEU}{InEU}
\DeclareMathOperator{\EUCountry}{EUCountry}
\DeclareMathOperator{\CountryInEU}{CountryInEU}
\DeclareMathOperator{\Turtle}{Turtle}
\DeclareMathOperator{\Shell}{Shell}
\DeclareMathOperator{\CanSwim}{CanSwim}
\DeclareMathOperator{\Musician}{Musician}
\DeclareMathOperator{\Love}{Love}
\DeclareMathOperator{\music}{music}
\DeclareMathOperator{\Benefit}{Benefit}
\DeclareMathOperator{\Apple}{Apple}
\DeclareMathOperator{\RedFruit}{RedFruit}
\DeclareMathOperator{\DirectedBy}{DirectedBy}
\DeclareMathOperator{\Filmmaker}{Filmmaker}
\DeclareMathOperator{\IsFilm}{IsFilm}
\DeclareMathOperator{\Singer}{Singer}
\DeclareMathOperator{\Writer}{Writer}
\DeclareMathOperator{\Building}{Building}
\DeclareMathOperator{\Tall}{Tall}

\DeclareMathOperator{\SummerCamp}{SummerCamp}
\DeclareMathOperator{\NoClass}{NoClass}
\DeclareMathOperator{\Entrepreneurs}{Entrepreneurs}
\DeclareMathOperator{\markZuckerberg}{markZuckerberg}
\DeclareMathOperator{\HateWorkingForOthers}{HateWorkingForOthers}
\DeclareMathOperator{\RiskAverse}{RiskAverse}
\DeclareMathOperator{\OlympicGoldMedalWinner}{OlympicGoldMedalWinner}
\DeclareMathOperator{\Athlete}{Athlete}
\DeclareMathOperator{\GoodAtSports}{GoodAtSports}
\DeclareMathOperator{\Fish}{Fish}
\DeclareMathOperator{\Sting}{Sting}
\DeclareMathOperator{\Youtube}{Youtube}
\DeclareMathOperator{\Instagram}{Instagram}
\DeclareMathOperator{\App}{App}
\DeclareMathOperator{\Attended}{Attended}
\DeclareMathOperator{\williamdickinson}{williamdickinson}
\DeclareMathOperator{\westminster}{westminster}
\DeclareMathOperator{\Highschool}{Highschool}
\DeclareMathOperator{\universityofedinburgh}{universityofedinburgh}
\DeclareMathOperator{\University}{University}
\DeclareMathOperator{\Be}{Be}
\DeclareMathOperator{\perfect}{perfect}
\DeclareMathOperator{\Play}{Play}
\DeclareMathOperator{\sam}{sam}
\DeclareMathOperator{\Song}{Song}
\DeclareMathOperator{\Poem}{Poem}
\DeclareMathOperator{\callus}{callus}
\DeclareMathOperator{\AgingAnalogy}{AgingAnalogy}
\DeclareMathOperator{\Debut}{Debut}
\DeclareMathOperator{\westworld}{westworld}
\DeclareMathOperator{\TVSeries}{TVSeries}
\DeclareMathOperator{\Adapt}{Adapt}
\DeclareMathOperator{\Produce}{Produce}
\DeclareMathOperator{\Write}{Write}
\DeclareMathOperator{\michael}{michael}
\DeclareMathOperator{\Direct}{Direct}
\DeclareMathOperator{\About}{About}
\DeclareMathOperator{\robots}{robots}

\newcommand{\stdf}[1]{\mbox{\scriptsize $\pm$#1}}

\usepackage[capitalise]{cleveref}

\title{Do LLMs Really Struggle at NL-FOL Translation? \\ Revealing their Strengths via a Novel Benchmarking Strategy
}
\author {
    Andrea Brunello\textsuperscript{\rm 1},
    Luca Geatti\textsuperscript{\rm 1},
    Michele Mignani\textsuperscript{\rm 1},
    Angelo Montanari\textsuperscript{\rm 1},
    Nicola Saccomanno\textsuperscript{\rm 1}\thanks{This work is the full version (appendix and code) of \citet{aaai2026}. Please, cite that when referring to this work.}
}
\affiliations {
    \textsuperscript{\rm 1}University of Udine, Italy\\
    name.surname@uniud.it
}

\begin{document}
\input{main}

\end{document}

%% file: main.tex
\maketitle

\begin{abstract}
    Due to its expressiveness and unambiguous nature, First-Order Logic (FOL) is a powerful formalism for representing concepts expressed in natural language (NL). This is useful, e.g., for specifying and verifying desired system properties.
    While translating FOL into human-readable English is relatively straightforward, the inverse problem, converting NL to FOL (NL-FOL translation), has remained a longstanding challenge, for both humans and machines. Although the emergence of Large Language Models (LLMs) promised a breakthrough, recent literature provides contrasting results on their ability to perform NL-FOL
    translation.
    In this work, we provide a threefold contribution.
    First, we critically examine existing datasets and protocols for evaluating NL-FOL translation performance, revealing key limitations that may cause a misrepresentation of LLMs' actual capabilities. 
    Second, to overcome these shortcomings, we propose a novel evaluation protocol explicitly designed to distinguish genuine semantic-level logical understanding from superficial pattern recognition, memorization, and dataset contamination. 
    Third, using this new approach, we show that state-of-the-art, dialogue-oriented LLMs demonstrate strong NL-FOL translation skills and a genuine grasp of sentence-level logic, whereas embedding-centric models perform markedly worse.

\end{abstract}


\begin{links}
    \link{Code}{https://github.com/dslab-uniud/NL-FOL-LT}
\end{links}

\section{Introduction}

Natural language (NL) stands as humanity's primary and most intuitive means of encoding and transmitting knowledge, thanks to its  expressiveness and remarkable flexibility. Nevertheless, these very characteristics, while enabling rich communication, also introduce challenges such as inherent ambiguity and often a lack of complete context.
To overcome these issues, formal languages, like First-Order Logic (FOL), offer a powerful alternative. FOL provides an unambiguous and highly expressive framework for representing complex concepts, making it particularly valuable for tasks like, for instance, specifying and verifying system properties. This is especially crucial in critical scenarios, where domain experts must completely and unambiguously define desired or unwanted system behaviour to perform formal verification. However, despite their usefulness, the correct specification and interpretation of logic formulas typically requires a strong mathematical background, severely limiting their applicability by untrained personnel \cite{barker2008empirical,barker2009difficulty,mpagouli2007converting}. 
In these contexts, a system capable of automatically translating between NL, like English, and logic would be of great help. 
Such a system would not only make formal methods more accessible but also find important applications in emerging domains, like AI Safety. For instance, it could enable the real-time monitoring of Large Language Models (LLMs)' behavior through automatic translation into logic of their outputs and internal reasoning (e.g., Chain-of-Thought monitoring \cite{korbak2025chain}), offering a key pathway to integrating symbolic and subsymbolic AI, or it could assist in the creation and in the online updating of formal world models and safety specifications 
\cite{Toward_guaranteed_safe_AI}.
Translating FOL into human-readable English is a relatively straightforward process, feasible even with a simple, attribute grammar-based formula parsing task \cite{ranta2011translating}. On the opposite, the problem of natural language to FOL (NL-FOL) translation, which can be considered as a subtask of autoformalization \cite{DBLP:conf/mkm/Szegedy20} and semantic parsing \cite{wilks1992preference}, has proven to be a persistent challenge for both human and artificial intelligence \cite{barker2009difficulty,singh2020exploring}. 
Despite recent advancements in natural language processing, the literature presents conflicting results on the ability of even state-of-the-art LLMs to achieve accurate and robust NL-FOL translation. Some works, like \citet{MALLS}, indicate good performance, while others, such as \citet{FOLIO}, show the opposite. These discrepancies arise because, often, studies perform evaluations on their own datasets with non-theoretically grounded protocols, hindering 
comparisons.

In this work, we aim to bring clarity to this situation, 
through three key contributions:
\begin{itemize}
\item We provide a critical assessment of the two most recent and comprehensive works which evaluate LLM performance in NL-FOL translation, highlighting issues with both the datasets and test protocols used, which may misrepresent LLMs' actual capabilities.
\item Based on our critique, we propose a novel, general, and theoretically grounded benchmarking strategy for NL-FOL translation. Departing from previous protocols, our approach restructures the formula generation task into two distinct phases and introduces additional subtasks designed to distinguish genuine logical understanding from superficial pattern recognition, memorization, and dataset contamination.
\item Finally, applying our benchmarking strategy, we show that the dialogue-oriented LLMs (OpenAI’s \textsc{GPT-4o-mini}, \textsc{o3-mini} and Qwen models \textsc{Qwen3-8B}, \textsc{Qwen3-30B-A3B}) achieve strong NL-FOL translation performance, with an authentic grasp of sentence-level logic. On the contrary, the state-of-the-art \textcolor{blue}, according to the MTEB leaderboard \cite{mteb_leaderboard}, embedding-centric models \textsc{Qwen3-Embedding-8B} and \textsc{Gemini-Embedding-001} perform markedly worse.  
\end{itemize}

The remainder of the paper is structured as follows. We first provide a summary of relevant literature, focusing on NL-FOL translation while also considering the more general task of autoformalization. We then present our contributions: a principled critique of the most recent and relevant evaluation protocols for NL-FOL translation, the detailed description of our novel benchmarking strategy, and 
the experimental results obtained from its application. Finally, we conclude and outline future research directions. In the \cref{app:fol}, 
we provide background on FOL.

\section{Related Work}

    In the literature \cite{DBLP:conf/mkm/Szegedy20,DBLP:conf/nips/WuJLRSJS22}, \emph{autoformalization} denotes the automatic translation of natural language statements into a given formalism. The focus has often been on formalizing mathematical proofs \cite{DBLP:conf/nips/WuJLRSJS22,ProofNET,DBLP:journals/corr/abs-2301-02195} so that they could be checked by interactive theorem provers such as Lean \cite{lean}, Isabelle \cite{isabelle}, and Coq \cite{Coq}.
    
    More recently, researchers have moved beyond the mathematical domain, motivated by the opportunity to use autoformalization as a bridge to combine symbolic and subsymbolic AI methodologies, or by the need to formally guarantee the safety of AI systems \cite{DBLP:conf/mkm/Szegedy20,DBLP:journals/cacm/SeshiaSS22,Toward_guaranteed_safe_AI}. Among the considered formalisms are Structured Query Language (SQL) \cite{DBLP:journals/elektrik/KanburogluT24}, Linear Temporal Logic (LTL) \cite{DBLP:conf/time/BrunelloMR19,DBLP:conf/fmcad/MendozaHT24}, and First-Order Logic (FOL).
    
    Regarding FOL, the translation from NL to this formalism has been addressed through various approaches, from rule-based methods \cite{DBLP:conf/emnlp/Abzianidze17,DBLP:conf/mlcw/BosM05,DBLP:conf/uai/ZettlemoyerC05}, to subsymbolic ones \cite{38d9186591d147e8ba3caec40080d1c4,DBLP:conf/acl/CaoZLLY19} and the use of language models like BERT and RoBERTa \cite{DBLP:conf/emnlp/TianLCX0J21}. With the advent of LLMs, several studies have investigated how to improve the initially modest performance of these models on the NL-FOL translation task \cite{DBLP:conf/coling/LuLGT0HXW22,MALLS,DBLP:journals/corr/abs-2409-16461}. 
    Recently, such a task has also been used instrumentally to create pipeline for enhancing general-purpose reasoning \cite{DBLP:conf/emnlp/PanAWW23,DBLP:conf/emnlp/OlaussonGLZSTL23,DBLP:conf/nips/YeCDD23} or automatic logical fallacies detection \cite{lalwani2025autoformalizingnaturallanguagefirstorder}. 
    The two most recent and comprehensive studies on NL-FOL translation are \citet{FOLIO} and \citet{MALLS}; for brevity, we refer to them as FOLIO and MALLS, respectively.
    Each introduces its own dataset and benchmarking protocol, yet they provide contradictory evidence about LLMs’ NL-FOL translation performance: FOLIO estimates roughly 52\% accuracy for \textsc{GPT-4} in a zero-shot setting (and 62\% for a few-shot), while, for the same model, MALLS seems to claim a much higher capability (around 80\% for their defined \emph{LE score} over their dataset). 
    In the following, we examine FOLIO and MALLS, 
    highlighting methodological shortcomings that may have led to a misinterpretation of LLMs’ logical competence in the NL-FOL translation task, and to the observed performance discrepancies.

\section{Limitations of Current Evaluation Protocols}
    
    To provide a precise analysis, we propose to first decompose the NL-FOL translation task into the following two steps:

    \smallskip

    \noindent \textbf{Ontology Extraction (OE)}: identify an appropriate
          logical signature (predicates, functions, constants), and associate to each
          logical symbol its intended semantic meaning.

    \smallskip

    \noindent \textbf{Logical Translation (LT)}: given the 
        signature, define 
        a FOL formula that captures the meaning of a 
        NL sentence.

    \smallskip
   
   This is crucial for multiple reasons: $(i)$ keeping OE and LT separate allows to determine whether a model fails at extracting a signature or at translating logic; $(ii)$ with a fixed, provided signature, LT verification is straightforward, e.g., a SMT solver~\cite{DBLP:reference/mc/BarrettT18} can compare the generated formula to the ground truth, but the same becomes much harder if each formula has its own symbols; $(iii)$ separating OE and LT helps identify techniques that work for one subtask but not the other, improving overall NL-FOL translation; $(iv)$ in many domains, experts can predefine an ontology once, so the system need only to perform logical translation over that fixed vocabulary; $(v)$ providing an ontology is also useful in all domains which require strict adherence to a given syntax.

By contrast, FOLIO and MALLS collapse OE and LT into a single task, an approach that introduces several evaluation problems, which we examine in the following.

    \subsection{Analysis of FOLIO \cite{FOLIO}}
        In FOLIO, the formalization task is considered alongside Natural Language Inference (NLI) to assess the logical understanding of models. The work comes with its own, expert-written dataset composed of 487 \emph{stories}, where a story is a list of NL \emph{premises} $p_1, \dots, p_n$ and their respective FOL translations $\varphi_1, \dots, \varphi_n$. Each story may repeat multiple (on average, three) times in the dataset, associated with a different \emph{conclusion} $c$. The latter is a phrase, paired with a formula $\psi$, and labeled with \textit{true}, \textit{false}, or \textit{unknown}, depending on whether $c$ is implied by the story, in contradiction with it, or cannot be resolved from the premises.
        
        For the NLI task, the model, given a story in NL, must determine the label of its (also NL) conclusion.
        
        For the formalization task, the model, given $p_1, \dots, p_n$ and $c$, must predict the logical translations $\varphi_1', \dots, \varphi_n'$ of the premises and the translation $\psi'$ of the conclusion: the formalization is considered correct if, depending on the label of $c$, $\varphi_1', \dots, \varphi_n'$ imply $\psi'$, or its negation, or neither. Note that this evaluation approach does not rely on the ground truth formulas $\varphi_1, \dots, \varphi_n, \psi$ of the premises and conclusion, and it works also with translations that use a different signature than the ground truth.\footnote{The ground truth formulas are in the dataset only to enable verification that the conclusion labels provided are correct.}
        However, it is a poor proxy for translation quality: it assigns the same score whether every sentence is mistranslated or only one is wrong, since in both situations the logical deduction step may fail.

        \subsection{Analysis of MALLS \cite{MALLS}}
        Unlike FOLIO, which judges translations via an SMT solver’s verdict, MALLS evaluates correctness by directly comparing each candidate translation to the respective ground truth formula. 
        Experiments are based on a synthetic dataset, generated by using OpenAI's GPT-4, and consisting of 28000 pairs of real-world NL statements and corresponding FOL translations.  
        Of these, 1000 have been human-verified and serve as the test set on which the paper’s results are based.

        The evaluation pipeline of MALLS attempts to estimate the similarity between the LLM prediction $\varphi'$ and the ground truth translation $\varphi$ using two scalar values: Logical Equivalence (LE) score, and BLEU score. While this method in principle enables a more precise evaluation, there are critical issues in how scores are attributed to the models' answers.

        \begin{table}[t]
            \centering
            \resizebox{0.8\linewidth}{!}{
            \begin{tabular}{ccccccccc}
            \toprule
            Formula & \multicolumn{8}{c}{Truth values}\\
            \midrule
                $\Country-Dummy$        & 0 & 0 & 0 & 0 & 1 & 1 & 1 & 1 \\
                $\InEU-\CountryInEU$    & 0 & 0 & 1 & 1 & 0 & 0 & 1 & 1 \\
                $\EUCountry-\EUCountry$ & 0 & 1 & 0 & 1 & 0 & 1 & 0 & 1 \\
                \midrule
                $\varphi$               & 1 & 1 & 1 & 1 & 1 & 1 & 0 & 1 \\
                $\varphi'$              & 1 & 1 & 0 & 1 & 1 & 1 & 0 & 1 \\
                \bottomrule
            \end{tabular}
            }
            \caption{Example of LE score calculation in MALLS.}
            \label{table_MALLS}                                         \end{table} 
        
        \paragraph{LE score is fundamentally flawed} To explain this metric and its limitations, 
        let us consider the following example, similar to the one reported in the original publication.
        Suppose we have a ground truth NL sentence $p : =$ \lq\lq Every country located in EU is an EU country\rq\rq, with ground truth translation  $\varphi:= \forall x \Country(x) \land \InEU(x) \to \EUCountry(x)$; and, let us consider the LLM prediction $\varphi' := \forall y \CountryInEU(y) \to \EUCountry(y)$.
        The LE score is computed in two stages. 
        First, a one-to-one mapping is established between the predicate symbols of the two formulas. In our example, $\InEU$ in $\varphi$ is paired with $\CountryInEU$ in $\varphi'$ and $\EUCountry$ in $\varphi$ with $\EUCountry$ in $\varphi'$; any extra predicate in one formula is matched to a fresh dummy symbol in the other  (here $\Country-Dummy$).
        Next, a truth table (see Table \ref{table_MALLS}) is generated. For every possible assignment of truth values to the predicates,  $\varphi$ and $\varphi'$ are evaluated by means of \emph{propositional logic} rules. The LE score is the fraction of columns in which the two formulas have identical truth values.
        Table~\ref{table_MALLS} clearly shows an issue with our 
        example: although $\varphi'$ can be regarded as a correct translation of $p$, it 
        gets a LE score below 1 ($7/8=0.875$). 
        
        Such a logical \lq\lq similarity\rq\rq{} computation is fundamentally flawed for two reasons: $(i)$ it treats FOL formulas $\varphi$ and $\varphi'$ as propositional logic ones, and assigning fixed truth values to predicate symbols is wrong, since the truth of a predicate may vary also with the object(s) it involves. For example, the LE score given to the pair \lq\lq$\exists x \ \Country(x) \ \land \ \InEU(x) \to \EUCountry(x)$\rq\rq{} and \lq\lq$\forall x \ \Country(x) \ \land \ \InEU(x) \to \EUCountry(x)$\rq\rq{} would be 1, despite their different meanings; $(ii)$  the columns of Table \ref{table_MALLS} can be viewed as world models, yet some of them are semantically non meaningful: any column that assigns 1 to $\EUCountry-\EUCountry$ while giving 0 to  $\Country-Dummy$ violates the obvious constraint that every EU country is a country. 
        The LE score treats the truth values of predicate symbols as independent, while, often they are interdependent.
        No simple, automatic fix seems to exist for this problem; for a related discussion in another setting, see \cite{Paris_Vencovská_2015}.

        \paragraph{BLEU score does not capture the logical semantic} Given two texts, BLEU \cite{DBLP:conf/acl/PapineniRWZ02} computes the geometric mean of the modified \(n\)-gram precisions for \(n = 1,\dots,4\), multiplied by a brevity penalty.  In MALLS it is applied to the formula strings \(\varphi\) and \(\varphi'\).
        However, because BLEU was designed for NL, it performs poorly on logical formulas: two expressions that differ only by a systematic predicate renaming, or that are logically but not syntactically equivalent, receive a low score.
        In the absence of tokenisation guidelines from the MALLS authors, if we treat every symbol in our 
        example’s \(\varphi\) and \(\varphi'\) as a separate token, we obtain an overall raw BLEU score of 0.18.

        \paragraph{MALLS Dataset contains ground truth errors due to misspecified guidelines used in the annotation process}
               
        Beyond flaws in the evaluation metrics, the pipeline used to human-verify the 1000 test instances also presents problems. The guidelines provided to annotators (Appendix A.3 of \cite{MALLS}) have at least two serious shortcomings.
        
        The first is about the handling of quantifiers. The guidelines state that $\forall$ (resp., $\exists$) should be used when the NL sentence explicitly contains \lq\lq  Every A \dots\rq\rq{} or \lq\lq For all A \dots\rq\rq{} (resp., \lq\lq Some A \dots\rq\rq{} or \lq\lq There exists A \dots\rq\rq{}), but \lq\lq if NL does not have these explicit hints, then using either $\forall$ or $\exists$ is fine\rq\rq. This creates errors. 
        Indeed, we find sentences like \lq\lq A child plays with a toy in a playground\rq\rq{} that are treated as universal assertions, and sentences like \lq\lq Birds can fly, while fish can swim, and elephants can neither fly nor swim\rq\rq{} that are translated with existential quantifications despite referring to entire subject classes. 
        Specifically, in the guidelines it is written that, for the sentence \lq\lq A turtle has a shell and can swim\rq\rq, a correct FOL formalization is $\exists x (\Turtle(x) \land \Shell(x) \land \CanSwim(x))$, while a more natural formula would be $\forall x (\Turtle(x) \to \Shell(x) \land \CanSwim(x))$.
        
        The second issue is about logical connectives interchangeability. The guidelines claim that $\to, \land$, and $\leftrightarrow$ are sometimes interchangeable. For instance, they report that also the following formulas are correct translations for the sentence above:
        \begin{gather*}
            \exists x (\Turtle(x) \to \Shell(x) \land \CanSwim(x)) \\ \exists x (\Turtle(x) \leftrightarrow \Shell(x) \land \CanSwim(x))
        \end{gather*}
        This is theoretically wrong and unacceptable for rigorous logical capability assessment and critical autoformalization applications. The first formula means that \lq\lq there is an entity that if it is a turtle, then it has a shell and it can swim\rq\rq, which can be true even if not all  turtles have a shell or can swim, and also in worlds in which there are no turtles; the second formula means that \lq\lq there is an entity that is a turtle if and only if it has a shell and it can swim\rq\rq.

\section{Our Novel Benchmarking Strategy}

    \begin{figure}[t]
        \centering
        \includegraphics[width=0.85\linewidth]{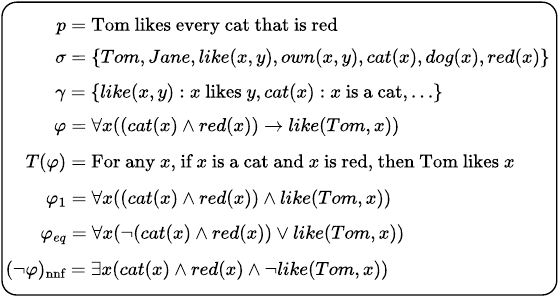}
        \caption{Example of instantiation of phrases, signatures, and formulas as used in the benchmarking strategy tasks.}
        \label{fig:bench_inst}
    \end{figure}

    Building upon the identified limitations of the existing evaluation protocols, in this section we introduce our novel benchmarking strategy to assess the proficiency of LLMs in the NL-FOL translation task.  
    The approach is outlined here in general terms, and the design choices are open to customization; concrete applications 
    on specific models and datasets are provided in the experimental section.
    
    Our method relies on a dataset $\mathcal{D}$ composed of triplets $(p,\varphi, \Omega)$, where $p$ represents a natural language utterance, 
    $\varphi$ is its corresponding FOL formula, and $\Omega$ is an ontology, i.e., a pair $(\sigma, \gamma)$, where $\sigma$ is the FOL signature (predicates, functions, constants) within which the formula $\varphi$ is written, and $\gamma$ is a natural language glossary that specifies the intended meaning of each symbol in $\sigma$. For an intuitive example of instantiation of phrases, ontologies and formulas as used in the following, please refer to Figure \ref{fig:bench_inst}.
    
    Given the dataset $\mathcal{D}$, to comprehensively evaluate LLM capabilities, we define three core tasks, referred to as: \emph{logical translation}, \emph{most similar}, and \emph{ranking}.

\paragraph{Logical translation}

    In this task, for each triplet $(p,\varphi, \Omega=(\sigma, \gamma))$, the LLM is given a prompt containing $p$ and $\Omega$ and must generate, using (a subset of) the symbols in $\sigma$ and their descriptions in $\gamma$, a formula $\varphi'$ that captures the meaning of $p$. We then automatically verify that $\varphi^\prime$ is logically equivalent to the reference formula $\varphi$ using an SMT solver. 
    The task succeeds if $\varphi^\prime$ is found to be equivalent to $\varphi$.
    
    Providing $\Omega$ marks a deliberate departure from earlier studies, which asked the model to derive $\varphi^\prime$ from $p$ alone. As discussed previously, by supplying the ontology up front, we disentangle the two stages of the NL-FOL translation pipeline, \emph{ontology extraction} and \emph{logical translation}, allowing us to isolate the latter and minimize confounding factors. 

    Observe that this task remains subject to other limitations: an LLM might have memorized the (public) dataset $\mathcal{D}$, or it could rely for the translation on purely syntactic heuristics, mechanically transforming $p$ into $\varphi^\prime$, without genuinely comprehending the underlying logical semantics. To address these issues, we designed two additional tasks that are fundamentally different and allow us to probe the model from different angles: \emph{most similar} and \emph{ranking}.

    \paragraph{Most similar}

    Given a triplet $(p,\varphi, \Omega=(\sigma, \gamma))$,  we produce up to $k$ random perturbations of $\varphi$.\footnote{Atomic propositions, for instance, admit fewer modifications.} Each perturbation is obtained by applying a single elementary edit to $\varphi$, which may include: replacing one Boolean connective (e.g., $\wedge$ with $\vee$); switching a quantifier (e.g., $\forall$ with $\exists$); inserting or removing a negation in front of a literal (in \cref{app:pert}, a discussion why we considered this kind of modifications).
    
    Putting together the original formula with its perturbations, we obtain the set $\mathcal{F}_{ms} = \{\varphi, \varphi_1, \dots, \varphi_k\}$. The model, given a prompt containing $p$ and the (shuffled) set $\mathcal{F}_{ms}$, must select the formula whose meaning is closest to $p$; we call this the \emph{FOL most similar} task.
    The task succeeds if $\varphi$ is selected.

    We also consider a parallel task in natural language. Let $T()$ be a translation function that converts a FOL formula into NL by substituting logical symbols and predicates with their glosses (details are in the \cref{app:transl}). The model now chooses the sentence that best matches $p$ from (the shuffled) $\mathcal{T}_{ms} = \{T(\varphi),T(\varphi_1),\dots,T(\varphi_k)\}$. We refer to this as the \emph{NL most similar} task.
    The task succeeds if $T(\varphi)$ is selected.

    \paragraph{Ranking}

    The \emph{most similar} task merely asks the model to pick the candidate whose meaning is closest to the original utterance.  Here we shift to a finer-grained setting. Given an integer $k$, for each triplet $(p,\varphi, \Omega=(\sigma, \gamma))$ we build the set $\mathcal{F}_{r} =\{\varphi,\ \varphi_1,\dots,\varphi_k,\ \lnot\varphi,\ (\lnot\varphi)_{\text{nnf}}, \varphi_{\text{eq}}\}$, where: $\varphi_1, \dots, \varphi_k$ are $k$ logical perturbations of $\varphi$, as in the \emph{most similar} task; $\lnot\varphi$ is the outright negation of $\varphi$; $(\lnot\varphi)_{\text{nnf}}$ is $\lnot\varphi$ rewritten in \emph{negation normal form}, with all negations pushed down to the predicate level, producing a formula that is syntactically distant from the original; and, $\varphi_{eq}$ is a formula logically equivalent to $\varphi$. 
    %
    To create $\varphi_{eq}$ we first select a subformula of $\varphi$ at random and then apply one randomly chosen, applicable transformation among: DeMorgan's Laws $\neg (\alpha \lor \beta) \equiv \neg \alpha \land \neg \beta$ or $\neg (\alpha \land \beta) \equiv \neg \alpha \lor \neg \beta$;  Double Negation Law $\alpha \equiv \neg( (\neg \alpha)_{\text{nnf}})$; Commutativity Law $\alpha \land \beta \equiv \beta \land \alpha$ or $\alpha \lor \beta \equiv \beta \lor \alpha$; Distributivity Law ($\alpha \land (\beta \lor \gamma) \equiv (\alpha \land \beta) \lor (\alpha \land \gamma)$ or $\alpha \lor (\beta \land \gamma) \equiv (\alpha \lor \beta) \land (\alpha \lor \gamma)$); and, Implication Expansion $\alpha \to \beta \equiv \neg \alpha \lor \beta$.

    We now evaluate the model on two ranking subtasks:
    \begin{itemize}
    \item \emph{FOL ranking}: given a prompt containing $p$ and the (shuffled) set of formulas $\mathcal{F}_{r}$, the model must order them from the most to the least semantically similar to the utterance $p$. 
    The task succeeds if: (i) $\varphi$ and $\varphi_{\text{eq}}$ are placed at the top positions of the rank; and (ii) $\lnot\varphi$, and $(\lnot\varphi)_{\text{nnf}}$ are placed at the bottom positions of the rank. Elements within the top and bottom may appear in any order. 
    
    \item \emph{NL ranking}: each formula is first rendered in English via the translation function $T()$, producing $\mathcal{T}_{r} = \{T(\varphi), T(\varphi_1), \dots, T(\varphi_k), T(\neg\varphi), T((\lnot\varphi)_{\text{nnf}}), T(\varphi_{\text{eq}})\}$ The model then ranks these (shuffled) sentences by their semantic proximity to $p$. 
    The task succeeds with conditions analogous to the \emph{FOL ranking} ones.
    \end{itemize}

Observe how the \emph{most similar} and \emph{ranking} tasks are far less vulnerable than the \emph{logical translation} one to typical LLM evaluation issues such as memorisation and dataset leakage, since their candidate sets $\mathcal{F}_{ms},\mathcal{T}_{ms},\mathcal{F}_{r},\mathcal{T}_{r}$ are generated on the fly and are not confined to the formulas and phrases in $\mathcal{D}$.
In addition, by requiring the model to select or order candidates by semantic proximity, these tasks probe finer-grained logical understanding rather than surface pattern matching; indeed, the closest match can be syntactically distant from the original, 
thereby discouraging reliance on purely syntactic or mechanical heuristics.

Finally, although our benchmarking strategy has been presented with dialogue-oriented LLMs in mind, it can be adapted to embedding-centric models. In that case we retain only the \emph{most similar} and \emph{ranking} tasks: the model generates 
an embedding for each candidate utterance or formula, and semantic proximity to the embedding of $p$ is measured with an embedding-distance metric, in our case, cosine similarity.

\subsection{Some words on Ontology Extraction}
In this work, we focus exclusively on the second step of our proposed NL--FOL evaluation pipeline, i.e., Logical Translation, and we assume access to a predefined and correct ontology~$\Omega$.
Ontology learning is a longstanding and extensively studied challenge in knowledge representation~\cite{ARMARY2025100693}. Nevertheless, unified evaluation metrics and standardized benchmarks for this task remain limited~\cite{du2024short}. 

Formulating direct approaches to OE evaluation warrants a dedicated and rigorous analysis comparable to the one we provide for LT in this paper. Alternatively, because the quality of an ontology depends heavily on both its intended application and its underlying formalism, a common evaluation strategy is to assess it indirectly by measuring its effectiveness within a relevant proxy task~\cite{Dellschaft2006, du2024short}. In our setting, such an indirect evaluation would still require first assessing a model’s capabilities in Logical Translation.

For these reasons, we do not address OE here and leave this subtask to future work. Completing this component will provide the full implementation of the proposed NL--FOL evaluation pipeline.

\section{Experiments}

    In this section we first introduce the concrete models and datasets over which we applied our benchmarking strategy. Then, we outline the experimental workflow. See \cref{app:computing} for our computing infrastructure.

    \subsection{Considered Models}

    We evaluate six models in total. Four dialogue-oriented models: \textsc{GPT-4o-mini}~\cite{4o_mini}, \textsc{o3-mini}~\cite{o3_mini} \textsc{Qwen3-8B}, and \textsc{Qwen3-30B-A3B} (hereafter \textsc{Qwen3-30B}) \cite{DBLP:journals/corr/abs-2505-09388}; and two embedding-centric models, \textsc{Qwen3-Embedding-8B} (run locally; hereafter \textsc{Qwen-Emb})~\cite{qwen3embedding} and \textsc{Gemini-Embedding-001} (accessed via Google Cloud; hereafter \textsc{Gemini-Emb})~\cite{lee2025geminiembeddinggeneralizableembeddings}.
    
    We selected the dialogue-oriented models to cover two points on the cost-capacity trade-off curve: \textsc{o3-mini} offers in principle the highest capacity while remaining affordable for large-scale API experiments \cite{o3_mini_bench}, whereas \textsc{GPT-4o-mini} trades some performance for a lower usage cost; similar remarks hold for the two models \textsc{Qwen3-8B} and \textsc{Qwen3-30B}. Note that, for both Qwen models, the results shown in the paper refer to their \emph{thinking} mode \cite{DBLP:journals/corr/abs-2505-09388}; an analysis of the performances of their \emph{not-thinking} counterparts is provided in the \cref{app:qwen_thinking}.
    
    As for the embedding-centric models, they are purpose-built for producing high-quality sentence embeddings, they are widely used in retrieval, semantic-similarity, and reranking pipelines, and they represent the state-of-the-art in that 
    paradigm as of July 2025 \cite{mteb_leaderboard}.
    The selected Qwen models enable a more detailed comparative analysis between dialogue-oriented and embedding-centric architectures within the same model family, as they are explicitly built upon a shared foundational model. Specifically, \textsc{Qwen3-Embedding-8B} and \textsc{Qwen3-8B} are both deriving (i.e., different finetunes) from \textsc{Qwen3-8B-Base}. 
    We did not perform an equivalent analysis for the Gemini and OpenAI families, due to the lack of publicly documented evidence clearly identifying which, if any, dialogue-based models correspond directly to the embedding-oriented models considered in this study, and the high computational costs such an investigation would entail.
    
    To prevent data leakage, 
    server-side training was disabled for OpenAI's models via the provided data controls settings. All Qwen models were run entirely on local hardware, and Google Cloud explicitly states that customer data are not used for training models \cite{google2023genaiprivacy}.

    \subsection{Considered Datasets}

    We make use of two datasets, 
    $\mathcal{D}_{\text{Stanford}}$ and $\mathcal{D}_{\text{FOLIO}}$. Here, we describe their content and the preprocessing applied to each.

    \paragraph{$\mathcal{D}_{\text{Stanford}}$} 
    The dataset is a 2001–2010 subset of the \textit{Grade Grinder Corpus Release 1.0} \cite{barker2011student}. Its files have never been made public and were kindly shared with us by the original authors. The corpus records students’ submissions, which can be both correct and incorrect, entered via the Grade Grinder tutoring platform while they worked through exercises in \emph{Language, Proof and Logic} \cite{LPL}. In each exercise, a natural-language description of a scene in Tarski’s World \cite{barker2007tarski} is provided, and students must supply a FOL formula that expresses the same content.
    Our extract consists of 159 high-quality instances. In the notation of our benchmark, each instance is composed of an utterance $p$ paired with its validated FOL translation $\varphi$. We manually defined an ontology $\Omega = (\sigma,\gamma)$ which applies indistinctly to every text-formula item. The \cref{app:Stanford} provides the full ontology and presents handcrafted instances that are inspired by the dataset but do not reproduce any of its private content.

    \paragraph{$\mathcal{D}_{\text{FOLIO}}$} 
    The dataset released with \cite{FOLIO} contains 487 \emph{stories}, split 70\%-15\%-15\% into training, validation, and test sets. As we mentioned, a story comprises several  \emph{premises} pairs (a premise is an utterance $p$ with its FOL translation $\varphi$) and, on average, is associated to three \emph{conclusions}. For the purposes of our study we use only the training portion (339 stories), focusing solely on the premises and ignoring the conclusions.
    For each story we manually defined a distinct ontology $\Omega=(\sigma,\gamma)$, and attached it to all $(p,\varphi)$ premise pairs in that story.
    Flattening the premises across stories then yielded 1667 $(p,\varphi,\Omega)$ triples. As a final step, to keep the dataset comparable with $\mathcal{D}_\text{Stanford}$, whose formulas never employ the XOR operator, we discarded any triplet whose $\varphi$ contained XOR, leaving us with 1565 instances.

    \medskip

    We excluded the MALLS \cite{MALLS} test set from our evaluation because, as noted earlier, its validation pipeline has serious flaws. We also omitted 
    the synthetic LogicNLI dataset \cite{DBLP:conf/emnlp/TianLCX0J21} since, according to the FOLIO authors \cite{FOLIO}, its FOL translations exhibit limited complexity and syntactical variation. We are unaware of other well-founded, non-synthetic NL-FOL datasets, challenging enough to serve as a benchmark.

    \subsection{Experimental Workflow}

    We conducted separate experimental workflows for the dialogue-oriented and embedding-centric models, which were applied on both $\mathcal{D}_\text{Stanford}$ and $\mathcal{D}_\text{FOLIO}$ datasets.
    For the dialogue models we followed OpenAI’s reproducibility guidelines \cite{openai2025advancedusage}. Each prompt was issued five times, changing only the seed parameter in the API call while keeping all other settings fixed.\footnote{Although seeding reduces variance, OpenAI notes that some residual nondeterminism may remain.}
    For the embedding models (\textsc{Qwen-Emb} and \textsc{Gemini-Emb}), since they produce deterministic vectors, each query was run once.
    The \cref{app:hyper} lists all hyper-parameter values and provides the 
    prompts used with the 
    models. 
    All code required to replicate our results is publicly available (see \cref{app:data_availability}).

    \subsubsection{Dialogue-oriented Models}

        \hphantom{}
        \smallskip

        We used zero-shot, Automatic Chain of Thought Prompting \cite{zhang2022automaticchainthoughtprompting} with all the models. Details of the prompts are provided in the \cref{app:prompts}.

        
        \paragraph{Logical translation}
        For each triplet $(p,\varphi,\Omega=(\sigma,\gamma))$ in a dataset, we built a system prompt that: defines the NL-FOL translation task, supplies the ontology $\Omega$, and specifies the required output format. The accompanying user prompt contains only the utterance $p$. We parsed the model’s reply with a Python script to extract the candidate formula $\varphi^\prime$ and checked its logical equivalence to the reference formula $\varphi$ with the Z3 solver \cite{de2008z3}. If the formulas were found to be equivalent we assigned a score of 1, otherwise 0. Overall performance was computed as the average of these scores over the dataset.
    
        \paragraph{Most similar}
        Considering \emph{FOL most similar}, given a triplet $(p,\varphi,\Omega=(\sigma,\gamma))$, we prepared a system prompt describing the task to be performed in general terms; in the user prompt we provided the reference $p$ and the (randomly shuffled) set $\mathcal{F}_{ms}$ containing the original FOL formula and $k$=8 perturbations.
        The output of the model was constrained, by means of the Structured Outputs feature of OpenAI API, to be an integer which, according to the model, is the position of the formula that most closely represents the meaning of $p$ in the set. 
        If the model selected the position in which $\varphi$ occurs, then it scores 1, otherwise 0. Overall performance was computed as the average of these scores over the dataset.
    
        \paragraph{Ranking}
        Considering the \emph{FOL Ranking}, given a triplet $(p,\varphi,\Omega=(\sigma,\gamma))$, in a system prompt we described the task to be performed in general terms; in the user prompt we provided the reference $p$ and the (randomly shuffled) set $\mathcal{F}_{r}$ containing the original FOL formula $\varphi$, $k$=3 perturbations, an equivalent version of $\varphi$, and two versions $\neg \varphi, (\neg \varphi)_{\text{nnf}}$ of its negation. 
        The model's output was again constrained, in this case to be a list of integers which, according to the model, should represent the ranking of the elements in $\mathcal{F}_{r}$, ordered from the formula that shares the most similar meaning with $p$ down to the least similar. We then associate the following scores to the ranking, following the criteria defined in our benchmarking strategy:
        \begin{itemize}
            \item we assign 1 if in the ranking the first two positions are the ones of $\varphi$ and $\varphi_{\text{eq}}$ (no matter the order) and 0 otherwise: we define the \emph{Ranking-Equivalence score} as the average of these values over the dataset.
            \item we assign 1 if in the ranking the last two positions are the ones of $\neg \varphi$ and $(\neg \varphi)_{\text{nnf}}$ (no matter the order) and 0 otherwise: we define the \emph{ Ranking-Negation score} as the average of these values over the dataset.
            \item we assign 1 if both the conditions above are met and 0 otherwise: we define the \emph{Ranking-Both score} as the average of these values over the dataset.
        \end{itemize}

        \smallskip
       

    \subsubsection{Embedding-centric Models}

    \hphantom{}
    \smallskip

    As previously discussed, our evaluation strategy extends also to embedding models. We start by describing the configuration of the models employed in our experiments.


     For \textsc{Qwen-Emb}, other than the \lq\lq classical\rq\rq{} embedding generation, where a single input $q$ is encoded, the model can also accept a user-supplied instruction $I$ that is prepended to the input $q$ to tailor the embedding to a specific task \cite{qwen3embedding}. We test both settings: the version which embeds $q$ alone (denoted \textsc{Qwen-Emb-plain}), and the instruction-augmented one, embedding the concatenation of $I$ and $q$ (denoted \textsc{Qwen-Emb-inst}).

    
     Concerning \textsc{Gemini-Emb}, it is possible to task the model to generate embeddings optimized for a specific task, selecting one among a list of predefined task types. We chose \texttt{SEMANTIC\_SIMILARITY} and embedded only the input $q$.

     \smallskip
   
    We now describe the \emph{most similar} and \emph{ranking} tasks (recall that \emph{logical translation} is not performed here).

    \paragraph{Most Similar} Considering \emph{FOL Most Similar}, given a triplet $(p,\varphi,\Omega=(\sigma,\gamma))$, and having the set  $F_{ms} = \{\varphi, \varphi_1, \dots, \varphi_8\}$, we computed the embeddings $v_p, e, e_1, \dots, e_8$  of $p$, $\varphi, \varphi_1, \dots, \varphi_8$ respectively (possibly augmented with the task instruction for \textsc{Qwen-Emb-inst}). We assigned a score of 1 if the embedding $e$ (of $\varphi$) exhibited the highest cosine similarity to $v_p$ among all the other candidate embeddings, and 0 otherwise. Overall performance was computed as the average of these scores over the dataset.
    
    \paragraph{Ranking} Considering \emph{FOL Ranking}, given the triplet $(p,\varphi,\Omega=(\sigma,\gamma))$, and having the set $\mathcal{F}_{r} = \{\varphi, \varphi_1, \dots, \varphi_3, \neg \varphi, (\neg \varphi)_{\text{nnf}},  \varphi_{\text{eq}}\}$, we computed the embeddings $v_p, e, e_1, \dots, e_3, e_{\text{neg1}}, e_{\text{neg2}}, e_{\text{eq}}$ of $p, \varphi, \varphi_1, \dots, \varphi_3, \neg \varphi, (\neg \varphi)_{\text{nnf}}, \varphi_{\text{eq}}$ respectively (possibly augmented with the task instruction for \textsc{Qwen-Emb-inst}). Then, the Ranking-Equivalence score was determined as in the dialogue-oriented case, considering that $e$ and $e_{\text{eq}}$ should be the two vectors 
    most similar to $v_p$ among all the candidates; for the Ranking-Negation score, $e_{\text{neg1}},$ and $ e_{\text{neg2}}$ should be the two least similar; 
    for the Ranking-Both score, both the conditions outlined above should be satisfied.

    \smallskip

    
    For the \emph{NL most similar} and the \emph{NL ranking} tasks with both dialogue-oriented and embedding-centric models, the only difference is that we consider the sets $\mathcal{T}_{ms}$ and $\mathcal{T}_{r}$ in place of $\mathcal{F}_{ms}$ and $\mathcal{F}_{r}$, respectively.

    \section{Results and Discussion}

    \begin{table}[t]
    \centering
    \resizebox{\linewidth}{!}{%
    \begin{tabular}{lcccc}
    \toprule
    & \textsc{GPT-4o-mini} & \textsc{o3-mini} & \textsc{Qwen3-8B} & \textsc{Qwen3-30B} \\ \midrule
    $\mathcal{D}_{\text{Stanford}}$ & .84\stdf{.03} & .94\stdf{.00} & .84\stdf{.01} & .85\stdf{.01} \\
    $\mathcal{D}_{\text{FOLIO}}$    & .73\stdf{.01} & .80\stdf{.00} & .72\stdf{.00} & .74\stdf{.01} \\
    \bottomrule
    \end{tabular}
    }
    \caption{Results of the dialogue-oriented models, logical translation. Average $\pm$ standard deviation over 5 repetitions.\label{tab:lt}}
    \end{table}

    \begin{table}[t]

      \centering
      \resizebox{\linewidth}{!}{%
      \begin{tabular}{ll*{10}{c}}
        \toprule
        \multirow{2}{*}{Task} & \multirow{2}{*}{Dataset} &
        \multicolumn{2}{c}{\textsc{GPT-4o-mini}} &
        \multicolumn{2}{c}{\textsc{o3-mini}} &
        \multicolumn{2}{c}{\textsc{Qwen3-8B}} &
        \multicolumn{2}{c}{\textsc{Qwen3-30B}} \\
        \cmidrule(lr){3-4} \cmidrule(lr){5-6} \cmidrule(lr){7-8}
        \cmidrule(lr){9-10} 
          & & NL & FOL & NL & FOL & NL & FOL & NL & FOL \\
        \midrule
        Most similar          & $\mathcal{D}_{\text{Stanford}}$ & .89\stdf{.02} & .88\stdf{.02} & .99\stdf{.00} & 1.0\stdf{.00} & .95\stdf{.01} & .96\stdf{.00} & .97\stdf{.01} & .98\stdf{.01} \\
                              & $\mathcal{D}_{\text{FOLIO}}$    & .91\stdf{.00} & .89\stdf{.00} & .95\stdf{.00} & .95\stdf{.00} & .93\stdf{.00} & .92\stdf{.00} & .93\stdf{.01} & .94\stdf{.00} \\
        Ranking-Eq.   & $\mathcal{D}_{\text{Stanford}}$ & .50\stdf{.03} & .57\stdf{.01} & .93\stdf{.02} & .98\stdf{.00} & .71\stdf{.03} & .79\stdf{.03} & .80\stdf{.01} & .91\stdf{.01} \\
                              & $\mathcal{D}_{\text{FOLIO}}$   & .52\stdf{.01} & .64\stdf{.01} & .91\stdf{.01} & .94\stdf{.00} & .77\stdf{.02} & .85\stdf{.00} & .86\stdf{.01} & .90\stdf{.00} \\
        Ranking-Neg.      & $\mathcal{D}_{\text{Stanford}}$ & .49\stdf{.02} & .51\stdf{.02} & .88\stdf{.02} & .91\stdf{.01} & .52\stdf{.05} & .56\stdf{.01} & .63\stdf{.01} & .62\stdf{.03}\\
                              & $\mathcal{D}_{\text{FOLIO}}$   & .62\stdf{.01} & .62\stdf{.01} & .84\stdf{.00} & .85\stdf{.01} & .62\stdf{.02} & .64\stdf{.01} & .70\stdf{.00} & .72\stdf{.01}\\
        Ranking-Both          & $\mathcal{D}_{\text{Stanford}}$ & .28\stdf{.02} & .32\stdf{.02} & .82\stdf{.02} & .89\stdf{.02} & .39\stdf{.05} & .46\stdf{.02} & .53\stdf{.02} & .57\stdf{.04}\\
                              & $\mathcal{D}_{\text{FOLIO}}$   & .39\stdf{.01} & .44\stdf{.01} & .78\stdf{.01} & .82\stdf{.01} & .51\stdf{.02} & .57\stdf{.01} & .62\stdf{.01} & .66\stdf{.01} \\
        \bottomrule
      \end{tabular}}
      \caption{Results on Most similar and Ranking for dialogue-oriented models: average $\pm$ standard deviation over 5 reps.}
      \label{table_results_dialogue}
    \end{table}

    \begin{table}[t]
    \centering
      \resizebox{\linewidth}{!}{%
      \begin{tabular}{ll*{10}{c}}
        \toprule
        \multirow{2}{*}{Task} & \multirow{2}{*}{Dataset} &
        \multicolumn{2}{c}{\textsc{Qwen-Emb-plain}} &
        \multicolumn{2}{c}{\textsc{Qwen-Emb-inst}} &
        \multicolumn{2}{c}{\textsc{Gemini-Emb}} \\
        \cmidrule(lr){3-4} \cmidrule(lr){5-6} \cmidrule(lr){7-8}
          & & NL & FOL & NL & FOL & NL & FOL \\
        \midrule
        Most similar          & $\mathcal{D}_{\text{Stanford}}$ & .62 & .46 & .75 & .55 & .69 & .52 \\
                              & $\mathcal{D}_{\text{FOLIO}}$  & .58 & .44 & .82 & .76 & .65 & .49 \\
        Ranking-Eq.   & $\mathcal{D}_{\text{Stanford}}$ & .30 & .21 & .30 & .31 & .30 & .29 \\
                              & $\mathcal{D}_{\text{FOLIO}}$  & .40 & .21 & .42 & .41 & .38 & .35 \\
        Ranking-Neg.      & $\mathcal{D}_{\text{Stanford}}$ & .51 & .36 & .70 & .71 & .67 & .64 \\
                              & $\mathcal{D}_{\text{FOLIO}}$  & .58 & .44 & .76 & .86 & .69 & .71 \\
        Ranking-Both          & $\mathcal{D}_{\text{Stanford}}$ & .18 & .09 & .25 & .23 & .22 & .20 \\
                              & $\mathcal{D}_{\text{FOLIO}}$ & .30 & .13 & .34 & .37 & .34 & .32 \\
        \bottomrule
      \end{tabular}
      }
      \caption{Results on Most similar and Ranking for embedding-centric models.}
      \label{table_results_embedding}
    \end{table}

    Here, we present the results obtained with our benchmarking strategy, starting with the logical translation task, where only dialogue-oriented models are involved. 
    Because we performed no prompt engineering (e.g., we did not use any few-shot exemplars) the reported scores should be viewed as lower bounds on the models’ capabilities. A thorough analysis of prompting strategies, which is beyond the scope of the present paper, is left for future work.

    \paragraph{Logical Translation}
    Table~\ref{tab:lt} reports the overall performance of the models evaluated with our metric. 
    All the models perform well on the considered datasets. The weakest result is the .72 achieved by \textsc{Qwen3-8B} on $\mathcal{D}_{\text{FOLIO}}$, meaning that, nevertheless, in 72\% of the cases the model translates a NL sentence into a FOL formula equivalent to the reference. 
    %
    Aligning with our expectations, \textsc{o3-mini} consistently outperforms \textsc{GPT-4o-mini} by a large margin; the same doesn't hold when considering \textsc{Qwen3-30B} and \textsc{Qwen3-8B}, despite the difference in size.
    Interestingly, all models find $\mathcal{D}_{\text{Stanford}}$ \lq\lq easier\rq\rq{} than $\mathcal{D}_{\text{FOLIO}}$ ($.11$--$.14$ difference). 
    
    We find no clear evidence that this discrepancy arises from a lower intrinsic difficulty of $\mathcal{D}_{\text{Stanford}}$, which includes formulas that challenge even advanced undergraduates (see, e.g., \citet{brunello2025evaluating} as well as our \cref{app:Stanford} for more details). 
    %
    A more plausible explanation for the observed performance lies in the quality of the dataset. As reported by \citet{DBLP:conf/emnlp/OlaussonGLZSTL23}, approximately 11\% of the original validation set of $\mathcal{D}_{\text{FOLIO}}$ contains errors. Since our experiments relied on the training portion of $\mathcal{D}_{\text{FOLIO}}$, and the train–validation–test split was performed randomly, a comparable proportion of errors is likely present in our data. Given the prohibitive length of the dataset, a full manual inspection was infeasible; therefore, we focused on instances where both models \textsc{GPT-4o-mini} and \textsc{o3-mini} consistently produced outputs judged incorrect across all random seeds. Among the 302 such instances examined, 93 were found to have incorrect ground-truth labels, indicating a conservative lower bound of roughly 6\% mistakes in the whole training dataset. Here an example (story-id 8, instance no. 3) where the dataset is wrong: 
    \begin{flushleft}
    NL: ``Some musicians love music.'' \\
    FOL (dataset): $\exists x (\Musician(x) \to \Love(x, \music))$ \\
    \textsc{o3-mini}: $\exists x (\Musician(x) \land \Love(x, \music))$
    \end{flushleft}
    A more detailed analysis of these cases is provided in the \cref{app:FOLIO}.

    \paragraph{Most similar}
    We now turn to the \emph{most similar} task, restricting 
    to dialogue-oriented models.
    Here, as reported in Table \ref{table_results_dialogue}, the performance gap between $\mathcal{D}_{\text{FOLIO}}$ and $\mathcal{D}_{\text{Stanford}}$ is smaller. This is reasonable, because the errors in $\mathcal{D}_{\text{FOLIO}}$ are less likely to influence the task outcome: a model may still select the correct formula $\varphi$ as the most similar to sentence $p$ if the perturbations in $\mathcal{F}_{ms}$ differ substantially in meaning. Errors may nevertheless still be affecting the evaluation, given the drop in \textsc{o3-mini} performance from $1.00$ on $\mathcal{D}_{\text{Stanford}}$ to $0.95$ on $\mathcal{D}_{\text{FOLIO}}$. In future work, we plan to more thoroughly investigate this aspect.
    Notably, the dialogue-oriented models achieve the same results on the \emph{most similar} task with both its FOL and NL variants; the absence of a 
    performance gap when dealing with formulas versus NL utterances may suggest that the models can reliably interpret the semantic content of FOL expressions.

    \paragraph{Ranking}
    The \emph{ranking} task appears generally more challenging than \emph{most similar}. A closer analysis reveals that, in fact, the latter is for the most part a subset of the former. For example, with \textsc{GPT-4o-mini} on $\mathcal{D}_{\text{FOLIO}}$ (FOL), the model succeeds in \emph{most similar} but fails in \emph{ranking-equivalence} in 31\% of the instances, whereas the opposite 
    occurs in only 4\% of the cases. This trend holds consistently across models and datasets, as detailed in the \cref{app:qualitative}.
    
    The performance gap between smaller and bigger models in the same family is significant: the latter ones consistently achieves markedly higher scores. A second pattern also emerges: models perform better on the FOL variant than on the NL one. The \textit{most similar} task was not sufficiently challenging to expose such differences, remarking the importance of carefully designing evaluation tasks with enough discriminative power to probe the models’ actual logical capabilities.
    Interestingly, \textit{Ranking-Equivalence} proves generally easier than \textit{Ranking-Negation}. A closer inspection reveals that the models misrank $\lnot \varphi$ and $(\lnot \varphi)_{\text{nnf}}$ at similar rates, despite the fact that $(\lnot \varphi)_{\text{nnf}}$ is, in principle, syntactically more distant from $\varphi$ (and from $\lnot \varphi$ itself). This finding offers additional evidence that the models do not rely on purely syntactic information to perform the ranking task.

    \medskip

    Overall, our results indicate that the strong performance in \emph{logic translation} (particularly by \textsc{o3-mini}) reflects a genuine grasp of the underlying logical content of both NL utterances and FOL formulas, as the models’ success on the \emph{most similar} and (for \textsc{o3-mini}) \emph{ranking} tasks cannot be convincingly explained by shallow pattern matching or incidental exposure to NL–FOL pairs from any leaked corpus.

    \paragraph{Embedding-centric models}
    We finally turn to the embedding-centric models.  As shown in Table~\ref{table_results_embedding}, they generally underperform the dialogue-oriented models considered, underlining the difference between the state-of-the-art of these two different paradigms of language models.
    Among the models considered, \textsc{Qwen-Emb-inst} achieves the highest scores, followed by \textsc{Gemini-Emb} and \textsc{Qwen-Emb-plain}: this suggests that supplying task-specific instructions when generating embeddings is indeed beneficial. 
    Interestingly, we can notice the embedding-centric models struggle far more with the tasks' FOL variants than with the NL counterparts (with the partial exception of \textsc{Qwen-Emb-inst}); this pattern is in contrast to what was observed for dialogue-oriented models. 
    Finally, again, the embedding-centric models find \emph{Ranking-Equivalence} easier than \emph{Ranking-Negation}.

    \medskip
   
   Overall, although our study focused on dialogue‐oriented models, these preliminary results show that our benchmarking strategy is also effective for embedding‐centric ones. Future work could include a deeper analysis, e.g., exploring alternative embedding‐similarity metrics.

    \begin{table}[t]
      \setlength{\tabcolsep}{4pt}

      \centering
      \resizebox{\linewidth}{!}{%
      \begin{tabular}{ll*{10}{c}}
        \toprule
            Model & Dataset &
        BLEU &
        LE &
        $r_{pb} (\text{BLEU},\text{Our})$ &
        $r_{pb} (\text{LE},\text{Our})$ \\
        \midrule
        
        \textsc{gpt-4o-mini}         
        & $\mathcal{D}_{\text{Stanford}}$ & .86\stdf{.08} & .93\stdf{.01} & .56 & .49 \\
        & $\mathcal{D}_{\text{FOLIO}}$    & .85\stdf{.07} & .91\stdf{.00} & .81 & .62 & \\
        
        \textsc{o3-mini}        
        & $\mathcal{D}_{\text{Stanford}}$ & .87\stdf{.06} & .94\stdf{.00} & .49 & .58 \\
        & $\mathcal{D}_{\text{FOLIO}}$    & .88\stdf{.05} & .92\stdf{.00} & .80 & .64 \\
        
        \textsc{Qwen3-8B}         
        & $\mathcal{D}_{\text{Stanford}}$ & .84\stdf{.09} & .93\stdf{.01} & .60 & .44 \\
        & $\mathcal{D}_{\text{FOLIO}}$    & .84\stdf{.01} & .92\stdf{.01} & .80 & .57 & \\

        \textsc{Qwen3-30B}         
        & $\mathcal{D}_{\text{Stanford}}$ & .83\stdf{.10} & .93\stdf{.00}  & .66 & .45\\
        & $\mathcal{D}_{\text{FOLIO}}$    & .84\stdf{.00} & .91\stdf{.01} & .83 & .63 & \\
                              
        \bottomrule
      \end{tabular}}
      \caption{Results on LT according to the BLEU and LE score, and their point biserial correlation ($r_{pb}$) with our metric.}
      \label{table_correlation}
    \end{table}

    \paragraph{Empirical evaluation of BLEU and LE score}
    Previously, we examined the theoretical limitations of the evaluation protocols adopted in MALLS. We now complement the discussion with a quantitative analysis of the shortcomings inherent in the metrics used. Table~\ref{table_correlation} reports the point-biserial correlations between BLEU or LE scores and the corresponding Z3 equivalent/not-equivalent outcome (our proposed correctness metric). Overall, the results show low, unstable, and dataset-dependent correlations (all with $p$-values $\ll .05$).
    For instance, for \textsc{o3-mini}, the correlation between BLEU and our metric averages .49 on $\mathcal{D}_{\text{Stanford}}$, but rises substantially to .80 on $\mathcal{D}_{\text{FOLIO}}$. Comparable fluctuations appear for LE and across other models, underscoring that traditional evaluation metrics are inconsistent and insufficient when compared to our theoretically grounded pipeline.

\section{Conclusions and Future Work}

    In this work we proposed a theoretically grounded benchmarking strategy for LLM-based NL-FOL translation that separates the task into Ontology Extraction and Logical Translation for fine-grained evaluation, while supplementing it with specific subtasks  less vulnerable to data leakage: this mitigates the risk to confound genuine logical understanding with pattern matching or memorisation. 
    We applied the benchmarking strategy to dialogue-oriented (GPT-4o-mini, o3-mini) and embedding-centric  (Qwen3-Embedding-8B, Gemini-Embedding-001) models, showing that the former consistently outperform the latter, and that their strong scores arise from semantic understanding rather than the use of  superficial syntactical regularities or memorised NL–FOL pairs.
    Our results remark the value of careful task design when assessing the logical capabilities of LLMs.

    We believe our work opens promising research directions. Other than the ones already mentioned in the paper, 
    a key 
    next step is to determine how difficult extracting an ontology is for current models and to develop techniques that can improve their performance.
    In addition, progress with NL-FOL translation requires high-quality, carefully curated datasets, possibly built (or validated) with the help of high-performing LLMs and explicitly accounting for polysemy. 
    Finally, we encourage the community to adopt and extend our benchmarking strategy. Experimenting with alternative prompts, adding evaluation subtasks, explicitly handling NL ambiguity, and introducing an even more fine-grained scoring has the potential to reveal new dimensions of LLMs' logical capabilities.


\section*{Acknowledgments}
    MM thanks Cristian Curaba for the valuable discussions and insights that contribute to this work.
    LG, AM, and NS acknowledge the
    support from the Interconnected Nord-Est Innovation Ecosystem (iNEST),
    which received funding from the European Union Next-GenerationEU (PIANO NAZIONALE DI RIPRESA E RESILIENZA (PNRR) – MISSIONE 4 COMPONENTE 2,
    INVESTIMENTO 1.5 – D.D. 1058 23/06/2022, ECS00000043).
    In addition, Angelo Montanari acknowledges the support from the MUR PNRR project FAIR - Future AI Research (PE00000013) also funded by the European Union Next-GenerationEU.
    This manuscript reflects only the authors’ views and opinions, neither the European Union nor the European Commission can be considered responsible for them.

\bibliography{aaai2026}



\appendix

\setcounter{secnumdepth}{1}

\section{First-Order Logic}
\label{app:fol}
In this section, we recall the background notions of First-Order Logic (FOL) that are behind the contents presented the main paper. We begin with the notion of first-order signature. 

\begin{definition}[First-order signature]
\label{def:signature}
  A first-order signature $\sigma$ is a set $\V \cup \C \cup \F \cup \R$
  such that:
  \begin{itemize}
    \item $\V$ is a (countably infinite) set of \emph{variables};
    \item $\C$ is a set of \emph{constant symbols};
    \item $\F$ is a set of \emph{function symbols}, each with its arity;
    \item $\R$ is a set of \emph{relation symbols}, each with its arity;
  \end{itemize}
  and $\V \cap \C \cap \F \cap \R = \emptyset$.
\end{definition}

The set of \emph{terms} in FOL is defined inductively as follows:
\begin{enumerate*}[label=(\roman*)]
  \item every variable in $\V$ and every constant in $\C$ is a term;
  \item if $t_1,\dots,t_n$ are terms and $f \in \F$ is a function symbol
    with arity $n$, then $f(t_1,\dots,t_n)$ is a term.
\end{enumerate*}

\begin{definition}[Syntax of FOL]
  An \emph{atomic formula} over the signature $\sigma = \V \cup \C \cup \F
  \cup \R$ is a string of type $p(t_1,\dots,t_n)$ such that $p \in \R$ is
  a relation symbol with arity $n$ and $t_i$ is a term, for all $i \in
  \{1,\dots,n\}$. The set of \emph{FOL formulas $\varphi$ over the signature
  $\sigma$} is inductively defined as follows:
  \begin{align*}
    \varphi \coloneqq \ 
         &p(t_1,\dots,t_n)           \choice \\
         &\lnot \varphi                 \choice
         \varphi \land \varphi            \choice
         \varphi \lor \varphi             \choice
         \varphi \to \varphi              \choice
         \varphi \leftrightarrow \varphi  \choice \\
         &\exists x \varphi    \choice
         \forall x  \varphi 
  \end{align*}
\end{definition}

A \emph{literal} is an atomic formula $p(t_1, \dots, t_n)$ (\emph{positive literal}) or its negation $\neg p(t_1, \dots, t_n)$ (\emph{negative literal}).

Among the logical symbols mentioned above, parentheses can also be used to clarify which subexpressions a logical connective applies to. For improved readability, parentheses are sometimes omitted; in such cases, the following operator precedence is assumed (from highest to lowest): $\{\lnot,\exists,\forall\}$,  $\land$,
$\lor$, $\to$, $\leftrightarrow$. 

A variable $x$ is said to be \emph{quantified} if it falls within the scope of a quantifier such as $\forall x$ (universal) or $\exists x$ (existential). Otherwise, it is considered a \emph{free variable}. A formula may contain both free and quantified variables: when a formula contains no free variables, it is called a \emph{closed formula} or a \emph{statement}.

We now define the semantics of FOL. The core of this definition is the notion of \emph{$\sigma$-structure}, for any signature $\sigma$.

\begin{definition}[$\sigma$-structure]
  Let $\sigma = \V \cup \C \cup \F \cup \R$ be a signature.
  A $\sigma$-structure $\mathfrak{A}$ is given by:
  \begin{itemize}
    \item a \emph{domain} $\D_{\mathfrak{A}} \neq \emptyset$;
    \item for all $c \in \C$, an element $c_{\mathfrak{A}} \in \D_{\mathfrak{A}}$;
    \item for all $f \in \F$ of arity $n$, a function $f_\mathfrak{A}
      : (\D_\mathfrak{A})^n \to \D_\mathfrak{A}$;
    \item for all $p \in \R$ or arity $n$, a set $p_\mathfrak{A} \subseteq
      (\D_\mathfrak{A})^n$.
  \end{itemize}
\end{definition}

Given a $\sigma$-structure $\mathfrak{A}$ and a variable evaluation $V : \V \to
\D_{\mathfrak{A}}$, for each term $t$, we define $\mathfrak{A}_V(t)$ inductively as follows:
\begin{enumerate*}[label=(\roman*)]
  \item if $t \in \V$, then $\mathfrak{A}_V(t) = V(t)$;
  \item if $t \in \C$, then $\mathfrak{A}_V(t) = t_\mathfrak{A}$;
  \item if $t = f(t_1,\dots,t_n)$ with $f \in \F$, then $\mathfrak{A}_V(t)
    = f_\mathfrak{A}(\mathfrak{A}_V(t_1),\dots,\mathfrak{A}_V(t_n))$.
\end{enumerate*}
In the following, we define the semantics of FOL.

\begin{definition}
  Let $\varphi$ be an FOL formula over the signature $\sigma$, let $\mathfrak{A}$
  be a $\sigma$-structure and let $V$ be a variable evaluation. We define
  the fact that \emph{$\mathfrak{A}_V$ satisfies $\varphi$}, denoted with $\mathfrak{A}_V
  \models \varphi$, inductively as follows:
  \begin{itemize}
    \item $\mathfrak{A}_V \models p(t_1,\dots,t_n)$ iff
      $(V(t_1),\dots,V(t_n)) \in p_\mathfrak{A}$, for all $p \in \R$;
    \item $\mathfrak{A}_V \models \lnot \varphi$ iff $\mathfrak{A}_V \not\models \varphi$;
    \item $\mathfrak{A}_V \models \varphi \land \varphi'$ iff $\mathfrak{A}_V \models \varphi$
      and $\mathfrak{A}_V \models \varphi'$;
    \item $\mathfrak{A}_V \models \varphi \lor \varphi'$ iff $\mathfrak{A}_V \models \varphi$ or
      $\mathfrak{A}_V \models \varphi'$;
    \item $\mathfrak{A}_V \models \varphi \to \varphi'$ iff $\mathfrak{A}_V
      \not\models \varphi$ or $\mathfrak{A}_V \models \varphi'$;
    \item $\mathfrak{A}_V \models \varphi \leftrightarrow \varphi'$ iff $\mathfrak{A}_V
      \models \varphi \to \varphi'$ and $\mathfrak{A}_V \models \varphi' \to \varphi$;
    \item $\mathfrak{A}_V \models \exists x \varphi$ iff there exists a $d
      \in \D_\mathfrak{A}$ such that $\mathfrak{A}_{V'} \models \varphi$, where $V'(x)
      = d$ and $V'(y) = V(y)$ for all $y \neq x$;
    \item $\mathfrak{A}_V \models \forall x  \varphi$ iff for all $d \in
      \D_\mathfrak{A}$ we have that $\mathfrak{A}_{V'} \models \varphi$, where $V'(x) = d$
      and $V'(y) = V(y)$ for all $y \neq x$.
  \end{itemize}
\end{definition}

We say that \emph{$\mathfrak{A}$ satisfies $\varphi$}, denoted with $\mathfrak{A} \models
\varphi$, iff $\mathfrak{A}_V \models \varphi$, for all variable evaluations $V$. Given two FOL formulas $\varphi$ and $\varphi'$ over the signature $\sigma$, we say that \emph{$\varphi$ is equivalent to $\varphi'$} when $\mathfrak{A}_V \models \varphi$ iff $\mathfrak{A}_V \models \varphi'$, for all $\sigma$-structures $\mathfrak{A}$ and all variable assignments $V$: this is the same to say that $\mathfrak{A}_V \models \varphi \leftrightarrow \varphi'$ for all $\sigma$-structures $\mathfrak{A}$ and all variable assignments $V$.

It is worth to notice that these notions, here presented in the general case, become clearer when $\varphi$ is a statement. In this case, since there are no free variables, the fact that $\mathfrak{A}_V \models \varphi$ doesn't depend on the variable assignment $V$. For any two statement $\varphi$ and $\varphi'$, they are \emph{equivalent} when for any $\sigma$-structure $\mathfrak{A}$, we have that $\mathfrak{A} \models \varphi$ iff $\mathfrak{A} \models \varphi'$, or equivalently, $\mathfrak{A} \models \varphi \leftrightarrow \varphi'$.

For every FOL formula $\varphi$, it is possible to transform it into an equivalent formula, over the same signature, such that the negation ($\lnot$) appears only in front of atomic formulas. This normal form is called \emph{Negation Normal Form} (NNF, for short). We define $(\cdot)_{\text{nnf}}$ as the function such that, given in input any FOL formula $\varphi$ it returns its equivalent formula in NNF.

To verify the equivalence between two formulas $\varphi$ and $\varphi'$ over the same signature $\sigma$, we employ SMT (\emph{Satisfiability Modulo Theories}) solvers~\cite{DBLP:reference/mc/BarrettT18}. These solvers are highly efficient tools for deciding the \emph{satisfiability} of first-order formulas; that is, they determine whether, given a first-order formula $\varphi$ over a signature $\sigma$ as input, there exists a $\sigma$-structure $\mathfrak{A}$ and a variable assignment $V$ such that $\mathfrak{A}_V \models \varphi$.

An SMT solver typically consists of two main components: a Boolean solver that handles the propositional structure of $\varphi$ (often based on the DPLL or CDCL algorithms, as in classical SAT solvers~\cite{DBLP:series/faia/336}) and a \emph{theory solver} that resolves conjunctions of formulas belonging to a specific theory, such as LRA (Linear Real Arithmetic).
Over the past decades, driven by the remarkable efficiency of SAT solvers, SMT solvers have evolved into powerful tools for automated reasoning, and they now play a crucial role in tasks such as formal verification and planning.
In our experiments, we used the SMT solver Z3~\cite{de2008z3}.

We reduce the problem of checking equivalence between two statements $\varphi$ and $\varphi'$ over the signature $\sigma$ to a satisfiability problem. Specifically, the statements $\varphi$ and $\varphi'$ are equivalent if and only if the statement $\lnot(\varphi \leftrightarrow \varphi')$ is not satisfiable. Formally, $\mathfrak{A} \models \varphi \leftrightarrow \varphi'$ for all $\sigma$-structures $\mathfrak{A}$ if and only if there does not exist a $\sigma$-structure $\mathfrak{A}$ such that $\mathfrak{A} \models \lnot(\varphi \leftrightarrow \varphi')$.

\section{On the Kind of $\varphi$ Perturbations Chosen}
\label{app:pert}

    A FOL formula can be modified in several ways. In the main paper, we discuss some logical perturbations that arise from changing a boolean connector with another, by replacing a quantifier with another, or inserting/removing a negation in front of a literal. Other than these modifications, we could also, for instance, have replaced a constant/relational symbol with another (preserving the original arity). We tried that, but in preliminary experiments we found that these types of changes are trivial to spot even for \textsc{GPT-4o-mini} in the \emph{most similar} task. Accordingly, we restricted our perturbations to those described in the paper, since they seem to provide a more challenging benchmark.

\section{Translation Function $T()$}
\label{app:transl}
    In the main paper, we discuss the use of the function $T()$ for our experiments, which is used to translate a FOL formula into an NL utterance: in this appendix, we give more details on how we have implemented $T()$, what are the choices we made, and what are the changes that can be done in further works.

    A straightforward manner to perform FOL to NL translation is a rule-based substitution scheme: replacing each symbol in a formula $\varphi$ with a fixed natural language rendering. 
    Note that there are various ways in which these translations can be made: for instance, we can translate $\forall$ with \lq\lq for any\rq\rq{} or with \lq\lq for all\rq\rq, $\to$ with \lq\lq imply\rq\rq{} or \lq\lq if \dots then \dots \rq\rq.
    We need also to specify how to deal with relational and constant symbols, in order to translate atomic formulas such as $\Cube(A)$ in $\mathcal{D}_{\text{Stanford}}$ or $\Love(x_1, x_2)$ in $\mathcal{D}_{\text{FOLIO}}$. For each relational symbol we associate a \emph{positive meaning}, i.e., a string in NL that describes the meaning of the predicate: for instance, the positive meaning of $\Cube(x_1)$ is \lq\lq $x_1$ is a cube\rq\rq, and of $\Love(x_1, x_2)$ is \lq\lq $x_1$ loves $x_2$\rq\rq. Then, to complete the translation, we replace the placeholders $x_1$ and $x_2$ with the meaning of the predicate arguments: the variable symbols are not modified, while to each constant symbol is associated a meaning (in this case to the constant $A$ is associated \lq\lq A\rq\rq). 
    For the sake of readibility, for the translation we provide also a \emph{negative meaning} for each predicate symbol: for instance, $\Love(x_1, x_2)$ has a negative meaning \lq\lq $x_1$ doesn't love $x_2$\rq\rq{} and this will be used to translate $\neg \Love(x_1, x_2)$. Note that we exclude the functional symbols in our description, since we don't have them in the signatures used.

    Once established how to translate the logical operators, the predicate and the constant symbols, one can easily define a translation function $T()$. To give a more precise idea, we define here inductively the translation function used in the paper:\begin{itemize}
        \item for any variable $x$, $T(x)$ is \lq\lq x\rq\rq; for any constant symbol $c$, its translation $T(c)$ is the meaning of $c$; 
        \item for any atomic formula $\varphi = p(t_1, \dots, t_n)$, its translation $T(\varphi)$ is given by the positive meaning of the predicate symbol $p$, in which we replace the placeholders $x_1, \dots, x_n$ with $T(t_1), \dots, T(t_n)$;
        \item for any negative literal $\varphi = \neg p(t_1, \dots, t_n)$, its translation $T(\varphi)$ is given by the negative meaning of the predicate symbol $p$, in which we replace the placeholders $x_1, \dots, x_n$ with $T(t_1), \dots, T(t_n)$;
        \item $T(\varphi \land \psi)$ is \lq\lq $T(\varphi)$ and $T(\psi)$\rq\rq;
        \item $T(\varphi \lor \psi)$ is \lq\lq $T(\varphi)$ or $T(\psi)$\rq\rq;
        \item $T(\varphi \to \psi)$ is \lq\lq if $T(\varphi)$, then $T(\psi)$\rq\rq;
        \item $T(\varphi \leftrightarrow \psi)$ is \lq\lq $T(\varphi)$ if and only if $T(\psi)$\rq\rq;
        \item $T(\neg  \varphi)$ is \lq\lq it's false that $T(\varphi)$\rq\rq, if $\varphi$ is not atomic;
        \item $T(\exists x \varphi)$ is \lq\lq there is x such that $T(\varphi)$\rq\rq;
        \item $T(\forall x \varphi)$ is \lq\lq for all x $T(\varphi)$\rq\rq.
    \end{itemize}
    We then remove superfluous spaces or parentheses that can be still present in the translation, we add a period at the end of the sentence and capitalize the first character. 

    Obviously, different choices could have been made both in how to translate the logical operators and in how to specify the translation process. For example, not using the negative meaning of the predicate symbols, or leaving the parentheses of the original FOL formulas within the NL translations to obtain an unambiguous utterance.

\section{Computing Infrastructure}
\label{app:computing}
We performed the local experiments on a Dell PowerEdge R750 running Red Hat Enterprise Linux 8.7 (Ootpa). The system is equipped with two Intel(R) Xeon(R) Platinum 8360Y CPUs at 2.40 GHz (72 physical cores in total), 512 GB of RAM, 4 TB of disk storage, and an NVIDIA A100 GPU with 80 GB of VRAM.

\section{Qwen: \emph{Thinking} and \emph{not thinking} mode}
\label{app:qwen_thinking}
    For the Qwen models \textsc{Qwen3-8B} and \textsc{Qwen3-30B-A3B}, we conducted experiments in both \emph{thinking} and \emph{non-thinking} modes, as described in \cite{DBLP:journals/corr/abs-2505-09388}. While the main paper reports results for the \emph{thinking} mode, in this section we provide a comparison with the \emph{non-thinking} counterpart. Since, in the \emph{Logical Translation} task, the two modes yield identical average accuracy, we focus here on the \emph{Most Similar} and \emph{Ranking} tasks; the results are reported in Table \ref{table_not_thinking}. Overall, the comparison suggests the following:
\begin{itemize}
\item For \textsc{Qwen3-8B}, the performance differences between the thinking and non-thinking modes are marginal. In several cases, the thinking mode even produces a slight decrease in accuracy (notably in \emph{Ranking-Equivalence} and \emph{Ranking-Both}).
\item For \textsc{Qwen3-30B-A3B}, in contrast, the thinking mode generally improves performance, yielding substantial gains in certain tasks, up to a 6 point increment in accuracy for \emph{Ranking-Equivalence} and \emph{Ranking-Both}.
\end{itemize}

    \begin{table}[t]
    \centering
      \resizebox{\linewidth}{!}{%
      \begin{tabular}{ll*{10}{c}}
        \toprule
        \multirow{2}{*}{Task} & \multirow{2}{*}{Dataset} &
        \multicolumn{2}{c}{\textsc{Qwen3-8B}} &
        \multicolumn{2}{c}{\textsc{Qwen3-30B}} \\
    \cmidrule(lr){3-4} \cmidrule(lr){5-6} 
          & & NL & FOL & NL & FOL  \\
        \midrule
        Most similar          & $\mathcal{D}_{\text{Stanford}}$ & .97 & .97 & .98 & .98 \\
                              & $\mathcal{D}_{\text{FOLIO}}$  & .94 & .94 & .94 & .93 \\
        Ranking-Equivalence   & $\mathcal{D}_{\text{Stanford}}$ & .69 & .79 & .75 & .84 \\
                              & $\mathcal{D}_{\text{FOLIO}}$  & .78 & .84 & .84 & .86\\
        Ranking-Negation      & $\mathcal{D}_{\text{Stanford}}$ & .53 & .62 & .58 & .56 \\
                              & $\mathcal{D}_{\text{FOLIO}}$  & .67 & .71 & .67 & .68 \\
        Ranking-Both          & $\mathcal{D}_{\text{Stanford}}$ & .39 & .51 & .46 & .50 \\
                              & $\mathcal{D}_{\text{FOLIO}}$ & .56 & .62 & .59 & .60  \\
        \bottomrule
      \end{tabular}
      }
      \caption{Results for \textsc{Qwen}'s family models in the \emph{not thinking} mode (average across the seeds)}
      \label{table_not_thinking}
    \end{table}

\section{Details about $\mathcal{D}_{\text{Stanford}}$ Dataset}
\label{app:Stanford}
    \begin{table*}[t]
        \centering
        \resizebox{\linewidth}{!}{
        \begin{tabular}{ll}
            \toprule
             NL sentence & FOL translation \\
             \midrule
             If D is a tetrahedron, then it's to the left of B if and only if it is also in front of C & $\Tet(D) \to (\LeftOf(D,B) \leftrightarrow \FrontOf(D,C))$ \\
              Every dodecaedron has a cube that is to its left but is neither in back of or in front of it& $\forall x (\Dodec(x) \to \exists y (\Cube(y) \land \LeftOf(y,x) \land \neg \FrontOf(y,x) \land \neg \BackOf(y,x))$ \\
             Neither A nor C is left to either C or B & $\neg \LeftOf(A,C) \land \neg \LeftOf(A,B) \land \neg \LeftOf(C,B) \land \neg \LeftOf(C,C)$ \\
              Some cube is not medium & $\exists x (\Cube(x) \land \neg \Medium(x))$\\
              F is larger or smaller than D only if it's small & $(\Larger(F,D) \lor \Smaller(F,D)) \to \Small(F)$ \\
             Every tetrahedron to the left of a dodecahedron is also larger than it & $\forall x \forall y ((\Tet(x) \land \Dodec(y) \land \LeftOf(x,y)) \to \Larger(x,y))$ \\
             \bottomrule
        \end{tabular}
        }
        \caption{Handcrafted examples inspired by randomly sampled instances of $\mathcal{D}_{\text{Stanford}}$.}
        \label{table_stanford}
    \end{table*}

    Each NL sentence in the dataset is a description of a configuration of the Tarski's World \cite{barker2007tarski}.  Hence, the predicate symbols used represent geometric and spatial relations. They include:
    \begin{itemize}
        \item Arity 1: \lq\lq Cube\rq\rq, \lq\lq Tet\rq\rq, \lq\lq Dodec\rq\rq, \lq\lq Small\rq\rq, \lq\lq Large\rq\rq, \lq\lq Medium\rq\rq;
        \item Arity 2: \lq\lq Smaller\rq\rq,  \lq\lq Larger\rq\rq,  \lq\lq LeftOf\rq\rq,  \lq\lq RightOf\rq\rq, \lq\lq BackOf\rq\rq,  \lq\lq FrontOf\rq\rq, \lq\lq SameRow\rq\rq,  \lq\lq SameCol\rq\rq, \lq\lq SameSize\rq\rq,  \lq\lq SameShape\rq\rq, \lq\lq Adjoins\rq\rq;
        \item Arity 3: \lq\lq Between\rq\rq.
    \end{itemize}
    
    As outlined in \cite{brunello2025evaluating}, the dataset is considered challenging also by advanced undergraduate students. Some handcrafted examples of NL--FOL pairs, inspired by randomly sampled instances from the dataset, are shown in Table~\ref{table_stanford}. 
    We have conducted some simple experiments to compare the (syntactic) complexity of the FOL instances in $\mathcal{D}_{\text{Stanford}}$ with those in $\mathcal{D}_{\text{FOLIO}}$, finding the following:
    \begin{itemize}
        \item Percentage of atomic sentences: $\mathcal{D}_{\text{Stanford}}$ 0\% (our extract does not include them), $\mathcal{D}_{\text{FOLIO}}$ 14.4\%.
        \item In $\mathcal{D}_{\text{Stanford}}$ 2.6 logical operators (boolean operators and quantifiers) are used on average within a FOL formula, while in $\mathcal{D}_{\text{FOLIO}}$ 1.8.
        \item The 48\% of $\mathcal{D}_{\text{Stanford}}$ instances have three or more operators, while in FOLIO only the 23\%; with four or more operators, we get 19\% in $\mathcal{D}_{\text{Stanford}}$, while only 3\% in $\mathcal{D}_{\text{FOLIO}}$; with five or more operators, we get 6\% in $\mathcal{D}_{\text{Stanford}}$ and 0.1\% in $\mathcal{D}_{\text{FOLIO}}$
        \item As for the quantifiers: there are 63\% instances in $\mathcal{D}_{\text{Stanford}}$ with one quantifier, and 64\% in $\mathcal{D}_{\text{FOLIO}}$; with two, 18\% in $\mathcal{D}_{\text{Stanford}}$ and 1\% in $\mathcal{D}_{\text{FOLIO}}$.
    \end{itemize}

    While these statistics attest to the logical complexity of the FOL translations in $\mathcal{D}_{\text{Stanford}}$ (and allow a comparison with $\mathcal{D}_{\text{FOLIO}}$), drawing parallel conclusions about the paired NL sentences is more difficult. One could argue that, despite the complex formulas, the dataset might still be relatively easy if the utterances make the intended logic sufficiently explicit. Quantifying such \lq\lq explicitness\rq\rq, however, is inherently challenging.

    Nonetheless, we believe that the performance of embedding-centric models (Table~\ref{table_results_embedding} in the main paper) provides some indirect evidence. If the reference sentence $p$ and its logical translation $\varphi$ were closely aligned also at the structural/syntax level (other than semantic one), the considered embedding models should perform well on the \emph{most similar} task, because the embeddings of $p$ and $T(\varphi)$ (or, maybe to a lesser extent, of $p$ and $\varphi$) would be expected to lie close to each other in embedding space. Instead, the modest performance of the embeddig-centric models on $\mathcal{D}_{\text{Stanford}}$ (less than or equal to $\mathcal{D}_{\text{FOLIO}}$'s), despite their success on NL similarity benchmarks, suggests this is not the case.

\section{Experiments (Hyper-)Parameters}
    \label{app:hyper}
    In this appendix we discuss the (hyper-)parameters that we have used for the experiments.

    Regarding the dialogue-based models, all the experiments we discuss in the main paper have been performed through the OpenAI Chat Completions API: in particular, all the (hyper-)parameters are left with their default values except the \emph{seed}, the \emph{max completion tokens}, and the \emph{response format}. The seed parameter is used for the sake of reproducibility and the values it can assume are 3, 12, 26, 85, and 107. We set the max completion tokens to 2500 for \textsc{GPT-4o-mini} and 10000 for \textsc{o3-mini}, i.e., the smallest limits that allowed us to obtain complete answers from the models. As already mentioned in the paper, we used the \emph{structured outputs} feature for all the experiments discussed: we require the model to output the reasoning (i.e., a string), and then the final answer, that is, depending on the specific task, a string (for \emph{logical translation}), an integer (for \emph{most similar}), or a list of integers (for \emph{ranking}).

\section{Prompts}
\label{app:prompts}
    In Template 1-6, we list the prompts we used for the various experiments and types of models. Some prompt templates include placeholders, delimited by \texttt{<\textbackslash} and \texttt{\textbackslash>}, which are replaced with specific values at inference time. 
    
    \subsection{Dialogue-oriented models}
    As a general rule for dialogue-oriented models, we placed the task specification (and any required output format) in the system prompt, and the instance-specific input in the user prompt. 
    Note that, for the \emph{most similar} and \emph{ranking} tasks, we provide here only the prompts for the NL version of the experiments: with minor changes it is possible to obtain the FOL versions. The full prompts can be found in the code of our experiments within the accompanying supplementary material.

\subsection{Embedding-centric models}
    In the case of \textsc{Qwen-Emb-inst}, if the embedding is about a FOL formula the instruction $I$ is given by the string \lq\lq Encode the first-order logic meaning of the following first-order formula: \rq\rq ; if we are dealing with a NL sentence, then we provide the string \lq\lq Encode the first-order logic meaning of the following natural-language sentence: \rq\rq .

\section{Wrong/Incomplete Translations in $\mathcal{D}_{\text{FOLIO}}$}
\label{app:FOLIO}
In the main article, we noted how \cite{DBLP:conf/emnlp/OlaussonGLZSTL23} reported that approximately 11\% of the original validation set of $\mathcal{D}_{\text{FOLIO}}$ contains errors. For our experiments, we chose to work with the training set instead, due to its larger number of instances. However, since the training, validation, and test sets were randomly sampled from the same pool \cite{FOLIO}, it is reasonable to assume that a similar error rate may affect the portion of the dataset we used. 
Thus, we conducted a preliminary assessment (intended to be extended in future work) of the training split by analyzing the instances where the models \textsc{GPT-4o-mini} and \textsc{o3-mini} consistently produced outputs judged incorrect across all random seeds.

In the following, we present selected cases from the analyzed instances in which the ground truth FOL formula incorrectly encodes the NL utterance, leading to \textsc{o3-mini} being wrongly judged as incorrect. 
All the relational symbols used by \textsc{o3-mini} are allowed in the translation process: 
\begin{enumerate}[leftmargin=2.7em,label=Ex~\arabic*:]
    \item story-id = 8, instance no. 3:\\ NL: \lq\lq Some musicians love music" \\ FOL: $\exists x (\Musician(x) \to \Love(x, \music))$, \\\textsc{o3-mini}: $\exists x (\Musician(x) \land \Love(x, \music))$;
    \item  story-id = 409, instance no. 5,\\ NL: \lq\lq If K is not beneficial to people and also not an apple, then it is a fruit with the color red.", \\FOL: $\neg (\Benefit(k) \land \Apple(k)) \to \RedFruit(k)$, \\ \textsc{o3-mini}: $(\neg \Benefit(k) \land \neg \Apple(k)) \to \RedFruit(k)$;
    \item  story-id = 68, instance no. 2,\\ NL: \lq\lq If a film is directedy by a person, the person is a filmmaker.",\\ FOL: $\forall x \exists y (\DirectedBy(y, x) \to \Filmmaker(x))$, \\\textsc{o3-mini}: $ \forall x \forall y ((\IsFilm(x) \land \DirectedBy(x,y)) \to \Filmmaker(y))$;
    \item  story-id = 118, instance no. 4, \\NL: \lq\lq A musician can be a singer or a writer.", \\FOL: $ \exists x (\Musician(x) \land \Singer(x)  \lor \Writer(x))$, \\ \textsc{o3-mini}: $  \forall x (\Musician(x) \to (\Singer(x)  \lor \Writer(x)))$;
   	\item  story-id = 284, instance no. 1, \\ NL: \lq\lq Each building is tall.", \\ FOL: $ \forall x (\Building(x) \to  \neg \Tall(x))$, \\ \textsc{o3-mini}: $  \forall x (\Building(x) \to \Tall(x))$;
    \item  story-id = 423, instance no. 4, \\NL: \lq\lq All students not in summer camp do not have class during the weekend.\rq\rq, \\ FOL: $ \forall x (\SummerCamp(x) \to \NoClass(x))$, \\ \textsc{o3-mini}: $ \forall x ( \neg \SummerCamp(x) \to \NoClass(x))$;
   	\item  story-id = 321, instance no. 5, \\ NL: \lq\lq If Mark Zuckerberg is neither an entrepreneur nor a person who hates working for others, then Mark Zuckerberg is not a risk-averse person.", \\ FOL: $ \neg \Entrepreneurs(\markZuckerberg)  \lor  \neg \HateWorkingForOthers(\markZuckerberg) \to  \neg \RiskAverse(\markZuckerberg))$, \\\textsc{o3-mini}: $( \neg \Entrepreneurs(\markZuckerberg) \land  \neg \HateWorkingForOthers(\markZuckerberg)) \to  \neg \RiskAverse(\markZuckerberg)$;
    \item  story-id = 346, instance no. 2,\\ NL: \lq\lq All Olympic gold medal winners are good athletes.", \\FOL: $ \forall x (\OlympicGoldMedalWinner(x) \to \Athlete(x))$, \\\textsc{o3-mini}: $ \forall x (\OlympicGoldMedalWinner(x) \to (\Athlete(x) \land \GoodAtSports(x)))$;
   	\item  story-id = 171, instance no. 1, \\ NL: \lq\lq Some fish may sting.", \\FOL: $ \exists x  \exists y (\Fish(x) \to \Sting(x,y))$, \\ \textsc{o3-mini}: $ \exists x (\Fish(x) \land  \exists y \Sting(x,y))$;
   	\item story-id = 477, instance no. 3, \\NL: \lq\lq An APP is either related to YouTube or Instagram.",\\ FOL: $ \forall x (\Youtube(x)  \lor \Instagram(x))$,  \\\textsc{o3-mini}: $ \forall x (\App(x) \to (\Youtube(x)  \lor \Instagram(x)))$;
    \item  story-id = 146, instance no. 3, \\ NL: \lq\lq There is a strong analogy of human aging in the poem Callus 4.", \\FOL: $\Poem(\callus4) \to \AgingAnalogy(\callus4)$ \\ \textsc{o3-mini}: $\Poem(\callus4) \land \AgingAnalogy(\callus4)$.
\end{enumerate} 

In addition to logical errors in the ground truth FOL formulas, we also observed cases of imprecise or incomplete translations, which can affect evaluation outcomes. Notably, such imprecision may have had limited impact on the original FOLIO evaluation pipeline, which assesses formalization correctness at a coarser level, namely, across full stories. However, this tolerance does not extend to our evaluation protocol, which operates at a finer granularity, judging translation quality at the level of individual sentences.

For example, given a set of premises $p_1, \dots, p_n$, the corresponding formalizations $\varphi_1, \dots, \varphi_n$ may collectively offer an adequate representation of the overall narrative. However, the individual mappings $p_i \mapsto \varphi_i$ might not be exhaustive on their own: information missing from, e.g., $\varphi_1$ could be implicitly captured in, e.g., $\varphi_2$, due to redundancy or overlap between the sentences, which share a same domain.

Below, we provide examples of instances where such imprecisions occur and \textsc{o3-mini}'s answer is wrongly judged incorrect. Also here, all predicate symbols used by \textsc{o3-mini} are permitted during the translation process.

\begin{enumerate}[leftmargin=2.7em,label=Ex~\arabic*:]
	\item  story-id = 226, instance no. 2,\\ NL: \lq\lq William Dickinson attended Westminster school for high school and then the University of Edinburgh.", \\FOL: $\Attended(\williamdickinson, \westminster) \land \Highschool(\westminster) \land \Attended(\williamdickinson, \universityofedinburgh)$, \textsc{o3-mini}: $\Attended(\williamdickinson, \westminster) \land \Highschool(\westminster) \land \Attended(\williamdickinson, \universityofedinburgh) \land \University(\universityofedinburgh)$;
	\item  story-id = 284, instance no. 1, \\ NL: \lq\lq If a song is not \lq\lq Perfect\rq\rq , Sam will never play it.", \\FOL: $ \forall x ( \neg \Be(x, \perfect) \to  \neg \Play(\sam, x))$, \textsc{o3-mini}: $ \forall x ((\Song(x) \land  \neg \Be(x, \perfect)) \to  \neg \Play(\sam, x))$;
	\item  story-id = 235, instance no. 2, \\ NL: \lq\lq In 2016, a new television series named Westworld debuted on HBO.\rq\rq , \\ FOL: $\Debut(\westworld, year2016)$,   \\ \textsc{o3-mini}: $\TVSeries(\westworld) \land \Debut(\westworld, year2016)$;
	\item  story-id = 235, instance no. 3,\\ NL: \lq\lq The TV series Westworld is adapted from the original film in 1973, which was written and directed by Michael Crichton.\rq\rq , \\ FOL: $ \exists x (\Adapt(\westworld, x) \land \Produce(x, year1973) \land \Write(\michael, x) \land \Direct(\michael, x))$ \\\textsc{o3-mini}: $\TVSeries(\westworld) \land  \exists f ( \Adapt(\westworld, f) \land \Produce(f, year1973) \land \Write(\michael, f) \land \Direct(\michael, f))$;
	\item  story-id = 235, instance no. 4, \\ NL: \lq\lq The 1973 film Westworld is about robots that malfunction and begin killing the human visitors.\rq\rq , \\FOL: $\About(\westworld, \robots)$, \\ \textsc{o3-mini}: $\Produce(\westworld, year1973) \land \About(\westworld, \robots)$.
\end{enumerate}

\section{Qualitative analysis}
\label{app:qualitative}
    \subsection{Relation between most similar and ranking tasks}
    As already stated in the main paper, \emph{most similar} is generally easier than \emph{ranking-equivalence}: Table \ref{table_results_ms_r} provides a quantitative evidence for the claim, showing in how many instances (micro-average, 5 seeds) a model succeeds in one task and not in the other. For the sake of conciseness, we consider only the OpenAI models.
    
  \begin{table*}[t]
      \centering
      \resizebox{0.75\linewidth}{!}{%
      \begin{tabular}{ll*{10}{c}}
        \toprule
        \multirow{2}{*}{Model} & \multirow{2}{*}{Dataset} &
        \multicolumn{2}{c}{\textsc{MS and not R}} &
        \multicolumn{2}{c}{\textsc{not MS and R}} &
        \multicolumn{2}{c}{\textsc{MS and R}} &
        \multicolumn{2}{c}{\textsc{not MS and not R}} \\
        \cmidrule(lr){3-4} \cmidrule(lr){5-6} \cmidrule(lr){7-8}
        \cmidrule(lr){9-10} 
          & & NL & FOL & NL & FOL & NL & FOL & NL & FOL \\
        \midrule
        \textsc{GPT-4o-mini}          & $\mathcal{D}_{\text{Stanford}}$ & .44 & .35 & .04 & .03 & .45 & .54 & .07 & .08 \\
                              & $\mathcal{D}_{\text{FOLIO}}$    & .42 & .31 & .04 & .04 & .49 & .59 & .05 & .06 \\
        \textsc{o3-mini}   & $\mathcal{D}_{\text{Stanford}}$ & .07 & .02 & .01 & .00 & .92 & .97 & .00 & .00 \\
                              & $\mathcal{D}_{\text{FOLIO}}$  & .06 & .03 & .02 & .02 & .90 & .93 & .03 & .03 \\
        \bottomrule
      \end{tabular}}
      \caption{Relation between the most similar (MS) and ranking (R) tasks for \textsc{GPT-4o-mini} and \textsc{o3-mini}. \textsc{X and not Y} means task \textsc{X} succeeded while task \textsc{Y} did not.}
      \label{table_results_ms_r}
    \end{table*}

    \subsection{Relation between models in the same family}
    Another relevant analysis concerns the relationships between models within the same family and across different families. Our observations can be summarized as follows:
    \begin{itemize}
    \item Within the OpenAI family, the largest model (\textsc{o3-mini}) not only outperforms the smallest one (\textsc{GPT-4o-mini}), but the set of instances on which the former fails is generally a subset of those on which the latter fails. In contrast, this pattern is not as clear within the Qwen family.
    \item The performances of \textsc{GPT-4o-mini} and \textsc{Qwen3-8B} are comparable. However, among the instances where a failure occurs, approximately half are shared failures and half are not. This suggests that there exists a subset of instances consistently difficult for both models, while each model also struggles on additional instances that the other does not.
    \end{itemize}

    \begin{table*}[t]
      \centering
      \resizebox{\linewidth}{!}{%
      \begin{tabular}{ll*{10}{c}}
        \toprule
        Mod1 & Mod2
        & Dataset &
        \textsc{Mod1 and Mod2} &
        \textsc{Mod1 and not Mod2} &
        \textsc{not Mod1 and Mod2} &
        \textsc{not Mod1 and not Mod2} \\
        
        \midrule
        \textsc{GPT-4o-mini} & \textsc{o3-mini}   & $\mathcal{D}_{\text{Stanford}}$ & .80 & .03 & .11 & .06  \\
                              & & $\mathcal{D}_{\text{FOLIO}}$    & .71 & .02 & .08 & .19\\
        \midrule
        \textsc{Qwen3-8B} & \textsc{Qwen3-30B-A3B}   & $\mathcal{D}_{\text{Stanford}}$ & .76 & .08 & .09 & .07 \\
                              & & $\mathcal{D}_{\text{FOLIO}}$    & .66 & .06 & .07 & .21\\
        \midrule
        \textsc{GPT-4o-mini} & \textsc{Qwen3-8B}   & $\mathcal{D}_{\text{Stanford}}$  & .77 & .06 & .08 & .09\\
                              & & $\mathcal{D}_{\text{FOLIO}}$    & .66 & .07 & .06 & .21\\
        
        \bottomrule
      \end{tabular}}
      \caption{Relation between different models in the Logical Translation task}
      \label{table_results_comparison}
    \end{table*}

\section{Data and Code Availability}
\label{app:data_availability}

We provide all the necessary code to allow interested researchers to reproduce our experiments.

\smallskip

\texttt{LINK:} \url{https://github.com/dslab-uniud/NL-FOL-LT}

\smallskip

For detailed instructions on executing the code and reproducing the experiments, please refer to the \texttt{readme.md} file located in the home folder.

Concerning data availability, the $\mathcal{D}_{\text{FOLIO}}$ dataset is accessible from following the original authors \cite{FOLIO}. As for $\mathcal{D}_{\text{Stanford}}$, it is not publicly available by decision of the original authors. Yet, it is possible to ask them access to the dataset for research-related purposes; nevertheless, we included a detailed description of its structure and provided handcrafted instance examples that are representative of the dataset’s content.

\renewcommand{\lstlistingname}{Template}
\begin{lstlisting}[caption=\emph{Logical translation} (system prompt),captionpos=b,label=lst:prompt_1,float=t,mathescape,frame=single,basicstyle=\scriptsize\selectfont,numbers=none]
You are an expert evaluator specializing in translating natural language sentences into a logical formalism. Your task is to formalize a given sentence in First Order Logic (FOL).

Instructions:
You can use the following symbols: 
-  logical symbols: $\forall$ (for all), $\exists$ (exists), $\to$ (implies), $\leftrightarrow$ (is equivalent to), $\land$ (and), $\lor$ (or), $\neg$ (not)
- predicate symbols: <\predicate_symbols\>
- contant symbols: <\constant_symbols\>
- variable symbols: x,y,z,...
- non logical symbols: parenthesis ()

Output Format:
Return the First Order Logic sentence that best represents the meaning of the given sentence.

Input Format:
Sentence: {sentence}

\end{lstlisting}
\renewcommand{\lstlistingname}{Listing}

\renewcommand{\lstlistingname}{Template}
\begin{lstlisting}[caption=\emph{Logical translation} (user prompt),captionpos=b,label=lst:prompt_2,float=t,mathescape,frame=single,basicstyle=\scriptsize\selectfont,numbers=none]
Sentence: <\sentence\>
\end{lstlisting}
\renewcommand{\lstlistingname}{Listing}

\renewcommand{\lstlistingname}{Template}
\begin{lstlisting}[caption=NL version of \emph{most similar} (system prompt),captionpos=b,label=lst:prompt_3,float=t,mathescape,frame=single,basicstyle=\scriptsize\selectfont,numbers=none]
You are an expert evaluator specializing in semantic similarity assessment. Your task is to identify the rephrasing that best preserves the original meaning of a given sentence.
Instructions:

- You will receive one original sentence followed by multiple rephrased versions
- Evaluate each rephrasing based solely on semantic/logic equivalence (meaning preservation)
- Ignore differences in grammar, syntax, word order, or writing style
- Select and return the rephrasing that most accurately conveys the same meaning as the original sentence

Evaluation Criteria:

- Prioritize semantic accuracy over grammatical correctness
- Focus on whether the logical meaning is the same or not

Output Format:
Return only the  number of the selected rephrasing that best matches the original sentence's meaning.

Input Format:
Sentence: {sentence}

Rephrasing 1: {rephrasing_1}
Rephrasing 2: {rephrasing_2}
...
Rephrasing n: {rephrasing_n}

\end{lstlisting}
\renewcommand{\lstlistingname}{Listing}

\renewcommand{\lstlistingname}{Template}
\begin{lstlisting}[caption=NL version of \emph{most similar} (user prompt), captionpos=b,label=lst:prompt_4,float=t,mathescape,frame=single,basicstyle=\scriptsize\selectfont,numbers=none]
Sentence: <\reference\>
<\list_of_sentences_in_$\mathcal{F}_{ms}$\>
\end{lstlisting}
\renewcommand{\lstlistingname}{Listing}

\renewcommand{\lstlistingname}{Template}
\begin{lstlisting}[caption=NL version of \emph{ranking} (system prompt),captionpos=b,label=lst:prompt_5,float=t,mathescape,frame=single,basicstyle=\scriptsize\selectfont,numbers=none]
You are an expert evaluator specializing in ranking some NL sentences according to their semantic similarity. You will be given a reference and other sentences; your task is to rank these sentences depending on whether they convey the same meaning of the reference or not.

Instructions:
- Ignore difference in grammar, syntax, style, or word order. You are only interested in the semantic behind the sentences. It is possible that two sentences have a different logical structure but they convey the same logical meaning and this is the only thing you have to focus on;
- If one of the sentences is equivalent to the reference, it should be ranked first;
- If one of the sentences is equivalent to the negation of the reference, i.e. if it has the opposite meaning of the reference, it should be ranked last.
- If two sentences are equivalent, they should be ranked in adjacent positions

Input Format:
Reference: {reference}
Sentence1: {sentence1}
Sentence2: {sentence2}
...
SentenceN : {sentenceN}

Output Format:
Return, after a reasoning stage, the numbers of the sentences in order, from the number of the sentence that is the most similar to the reference to the one that is the least. ([number_of_the_statement_ranked_first, number_of_the_statement_ranked_second, ... number_of_the_statement_ranked_last]).
\end{lstlisting}
\renewcommand{\lstlistingname}{Listing}

\renewcommand{\lstlistingname}{Template}
\begin{lstlisting}[caption=NL version of \emph{ranking} (user prompt), captionpos=b,label=lst:prompt_6,float=t,mathescape,frame=single,basicstyle=\scriptsize\selectfont,numbers=none]
Sentence: <\reference\>
<\list_of_sentences_in_$\mathcal{F}_{r}$\>
\end{lstlisting}
\renewcommand{\lstlistingname}{Listing}

%% file: aaai2026.bib
@inproceedings{aaai2026,
  title={Do LLMs Really Struggle at NL-FOL Translation?
Revealing Strengths via a Novel Benchmarking Strategy},
  author={Brunello, Andrea and Geatti, Luca and Mignani, Michele and Montanari, Angelo and Saccomanno, Nicola},
  booktitle={AAAI-26},
  note={Accepted for publication},
  year={2026},
  organization={The 40th Annual AAAI Conference on Artificial Intelligence}
}

@inproceedings{ranta2011translating,
  title={Translating between language and logic: {W}hat is easy and what is difficult},
  author={Ranta, Aarne},
  booktitle={ 23rd international conference on Automated deduction (CADE)},
  pages={5--25},
  year={2011},
  organization={Springer}
}

@article{wilks1992preference,
  title={The preference semantics family},
  author={Wilks, Yorick and Fass, Dann},
  journal={Computers \& Mathematics with Applications},
  volume={23},
  number={2-5},
  pages={205--221},
  year={1992},
  publisher={Elsevier}
}

@article{korbak2025chain,
  author       = {Tomek Korbak and
                  Mikita Balesni and
                  Elizabeth Barnes and
                  Yoshua Bengio and
                  Joe Benton and
                  others},
  title        = {Chain of Thought Monitorability: {A} New and Fragile Opportunity for
                  {AI} Safety},
  journal      = {CoRR},
  volume       = {abs/2507.11473},
  year         = {2025},
  url          = {https://doi.org/10.48550/arXiv.2507.11473},
  doi          = {10.48550/ARXIV.2507.11473},
  eprinttype    = {arXiv},
  eprint       = {2507.11473},
  bibsource    = {dblp computer science bibliography, https://dblp.org}
}

@inproceedings{de2008z3,
  title={{Z3}: {A}n efficient {SMT} solver},
  author={De Moura, Leonardo and Bj{\o}rner, Nikolaj},
  booktitle={ 14th International conference on Tools and Algorithms for the Construction and Analysis of Systems (TACAS)},
  pages={337--340},
  year={2008},
  organization={Springer}
}

@misc{mteb_leaderboard,
    author = {{Huggingface}},
    title = {{MTEB Leaderboard}},
    howpublished = {\url{https://huggingface.co/spaces/mteb/leaderboard}},
    year = {2025}, 
    note = {Accessed: 2025-07-28}
}

@article{qwen3embedding,
  author       = {Yanzhao Zhang and
                  Mingxin Li and
                  Dingkun Long and
                  others},
  title        = {Qwen3 Embedding: Advancing Text Embedding and Reranking Through Foundation
                  Models},
  journal      = {CoRR},
  volume       = {abs/2506.05176},
  year         = {2025},
  url          = {https://doi.org/10.48550/arXiv.2506.05176},
  doi          = {10.48550/ARXIV.2506.05176},
  eprinttype    = {arXiv},
  eprint       = {2506.05176},
  bibsource    = {dblp computer science bibliography, https://dblp.org}
}

@article{lee2025geminiembeddinggeneralizableembeddings,
  author       = {Jinhyuk Lee and
                  Feiyang Chen and
                  Sahil Dua and others},
  title        = {Gemini Embedding: Generalizable Embeddings from Gemini},
  journal      = {CoRR},
  volume       = {abs/2503.07891},
  year         = {2025},
  url          = {https://doi.org/10.48550/arXiv.2503.07891},
  doi          = {10.48550/ARXIV.2503.07891},
  eprinttype    = {arXiv},
  eprint       = {2503.07891},
  bibsource    = {dblp computer science bibliography, https://dblp.org}
}

@misc{4o_mini,
  author       = {{OpenAI}},
  title        = {GPT-4o-mini},
  howpublished = {\url{https://openai.com/index/gpt-4o-mini-advancing-cost-efficient-intelligence/}},
  year         = {2024},
  note         = {Accessed: 2025-07-20}
}

@misc{o3_mini,
  author       = {{OpenAI}},
  title        = {o3-mini System Card},
  howpublished = {\url{https://openai.com/index/o3-mini-system-card/}},
  year         = {2025},
  note         = {Accessed: 2025-07-20}
}

@misc{o3_mini_bench,
  author       = {{OpenAI}},
  title        = {o3-mini benchmark},
  howpublished = {\url{https://openai.com/it-IT/index/openai-o3-mini/}},
  year         = {2025},
  note         = {Accessed: 2025-07-20}
}

@book{LPL,
    author = {Barker-Plummer, David and Barwise, Jon and Etchemendy, John},
    title = {Language, {P}roof, and {L}ogic},
    year = {2011},
    isbn = {1575866323},
    publisher = {Center for the Study of Language and Information/SRI},
    edition = {2nd}
}

@inproceedings{barker2011student,
  title={Student translations of natural language into logic: {T}he {G}rade {G}rinder corpus release 1.0},
  author={Barker-Plummer, Dave and Cox, Richard and Dale, Robert},
  booktitle={ 4th international conference on educational data mining (EDM)},
  pages={51-60},
  year={2011}
}

@misc{barker2007tarski,
  title={Tarski’s {W}orld: {R}evised and Expanded Edition},
  author={Barker-Plummer, Dave and Barwise, Jon and Etchemendy, John and Liu, Albert},
  year={2007},
  publisher={Leland Stanford Junior University: {CSLI} Publications}
}

@misc{google2023genaiprivacy,
  author       = {{Google Cloud}},
  title        = {Generative {AI}, Privacy, and {G}oogle {C}loud},
  howpublished = {\url{https://services.google.com/fh/files/misc/genai_privacy_google_cloud.pdf}},
  year         = {2023},
  month        = {August},
  note         = {Accessed: 2025-07-22}
}

@misc{openai2025advancedusage,
  author       = {{OpenAI}},
  title        = {{OpenAI} Platform Documentation: Advanced Usage},
  howpublished = {\url{https://platform.openai.com/docs/advanced-usage}},
  year         = {2025},
  note         = {Accessed: 2025-07-22}
}

@inproceedings{barker2008empirical,
  title={An empirical study of errors in translating natural language into logic},
  author={Barker-Plummer, Dave and Cox, Richard and Dale, Robert and Etchemendy, John},
  booktitle={ Annual Meeting of the Cognitive Science Society},
  volume={30},
  year={2008}
}

@inproceedings{barker2009difficulty,
  title={Dimensions of Difficulty in Translating Natural Language into First-Order Logic},
  author={Dave Barker-Plummer and Richard J. Cox and Robert Dale},
  booktitle={Educational Data Mining},
  year={2009},
  url={https://api.semanticscholar.org/CorpusID:551970}
}

@inproceedings{mpagouli2007converting,
  title={Converting first order logic into natural language: {A} first level approach},
  author={Mpagouli, Aikaterini and others},
  booktitle={ 11th Panhellenic Conference on Informatics (PCI)},
  pages={517--526},
  year={2007}
}

@article{singh2020exploring,
  author       = {Hrituraj Singh and
                  Milan Aggarwal and
                  Balaji Krishnamurthy},
  title        = {Exploring Neural Models for Parsing Natural Language into First-Order
                  Logic},
  journal      = {CoRR},
  volume       = {abs/2002.06544},
  year         = {2020},
  url          = {https://arxiv.org/abs/2002.06544},
  eprinttype    = {arXiv},
  eprint       = {2002.06544},
  bibsource    = {dblp computer science bibliography, https://dblp.org}
}

@inproceedings{DBLP:conf/nips/WuJLRSJS22,
  author       = {Yuhuai Wu and
                  Albert Qiaochu Jiang and
                  Wenda Li and
                  others},
  title        = {Autoformalization with Large Language Models},
  booktitle    = { 36th Annual Conference on Neural Information Processing Systems (NeurIPS)},
  year         = {2022}
}

@inproceedings{DBLP:conf/mkm/Szegedy20,
  author       = {Christian Szegedy},
  title        = {A Promising Path Towards Autoformalization and General Artificial
                  Intelligence},
  booktitle    = { 13th International Conference on Intelligent Computer Mathematics (CICM)},
  series       = {Lecture Notes in Computer Science},
  volume       = {12236},
  pages        = {3--20},
  publisher    = {Springer},
  year         = {2020},
  doi          = {10.1007/978-3-030-53518-6\_1}
}

@inproceedings{brunello2025evaluating,
  title={Evaluating LLMs Capabilities at Natural Language to Logic Translation: A Preliminary Investigation},
  author={Brunello, Andrea and Ferrarese, Riccardo and Geatti, Luca and Marzano, Enrico and Montanari, Angelo and Saccomanno, Nicola and others},
  booktitle={ 7th International Workshop on Artificial Intelligence and fOrmal VERification, Logic, Automata, and sYnthesis (OVERLAY)},
  volume={3904},
  pages={103--110},
  year={2025},
  organization={CEUR-WS}
}

@inproceedings{zhang2022automaticchainthoughtprompting,
  author       = {Zhuosheng Zhang and
                  Aston Zhang and
                  Mu Li and
                  Alex Smola},
  title        = {Automatic Chain of Thought Prompting in Large Language Models},
  booktitle    = {The Eleventh International Conference on Learning Representations,
                  {ICLR} 2023, Kigali, Rwanda, May 1-5, 2023},
  publisher    = {OpenReview.net},
  year         = {2023},
  url          = {https://openreview.net/forum?id=5NTt8GFjUHkr},
  bibsource    = {dblp computer science bibliography, https://dblp.org}
}

@inproceedings{MALLS,
  author       = {Yuan Yang and
                  Siheng Xiong and
                  Ali Payani and
                  Ehsan Shareghi and
                  Faramarz Fekri},
  title        = {Harnessing the Power of Large Language Models for Natural Language to First-Order Logic Translation},
  booktitle    = { 62nd Annual Meeting of the Association for Computational
                  Linguistics (ACL)},
  pages        = {6942--6959},
  publisher    = {ACL},
  year         = {2024},
  doi          = {10.18653/V1/2024.ACL-LONG.375}
}

@inproceedings{FOLIO,
  author       = {Simeng Han and
                  Hailey Schoelkopf and
                  Yilun Zhao and
                  Zhenting Qi and
                  others},
  title        = {{FOLIO:} {N}atural Language Reasoning with First-Order Logic},
  booktitle    = { 2024 Conference on Empirical Methods in Natural Language Processing (EMNLP)},
  pages        = {22017--22031},
  publisher    = {Association for Computational Linguistics},
  year         = {2024},
  doi          = {10.18653/V1/2024.EMNLP-MAIN.1229}
}

@article{Toward_guaranteed_safe_AI,
  author       = {David Dalrymple and
                  Joar Skalse and
                  Yoshua Bengio and
                  Stuart Russell and
                  others},
  title        = {Towards Guaranteed Safe {AI:} {A} Framework for Ensuring Robust and Reliable {AI} Systems},
  journal      = {CoRR},
  volume       = {abs/2405.06624},
  year         = {2024},
  url          = {https://doi.org/10.48550/arXiv.2405.06624},
  doi          = {10.48550/ARXIV.2405.06624}
}

@inproceedings{lean,
  author       = {Leonardo Mendon{\c{c}}a {De Moura} and
                  Soonho Kong and
                  Jeremy Avigad and 
                                  others},
  title        = {The Lean Theorem Prover},
  booktitle    = { 25th International Conference on
                  Automated Deduction (CADE)},
  series       = {Lecture Notes in Computer Science},
  volume       = {9195},
  pages        = {378--388},
  publisher    = {Springer},
  year         = {2015},
  url          = {https://doi.org/10.1007/978-3-319-21401-6\_26},
  doi          = {10.1007/978-3-319-21401-6\_26}
}

@book{isabelle,
  author       = {Lawrence C. Paulson},
  title        = {Isabelle - {A} Generic Theorem Prover},
  series       = {Lecture Notes in Computer Science},
  volume       = {828},
  publisher    = {Springer},
  year         = {1994},
  url          = {https://doi.org/10.1007/BFb0030541},
  doi          = {10.1007/BFB0030541},
  isbn         = {3-540-58244-4}
}

@misc{lalwani2025autoformalizingnaturallanguagefirstorder,
      title={Autoformalizing Natural Language to First-Order Logic: {A} Case Study in Logical Fallacy Detection}, 
      author={Abhinav Lalwani and Tasha Kim and Lovish Chopra and others},
      year={2025},
      eprint={2405.02318},
      archivePrefix={arXiv},
      primaryClass={cs.CL},
      url={https://arxiv.org/abs/2405.02318}, 
}

@article{ProofNET,
  author       = {Zhangir Azerbayev and
                  Bartosz Piotrowski and
                  Hailey Schoelkopf and
                  others},
  title        = {{ProofNet}: {A}utoformalizing and Formally Proving Undergraduate-Level
                  Mathematics},
  journal      = {CoRR},
  volume       = {abs/2302.12433},
  year         = {2023},
  url          = {https://doi.org/10.48550/arXiv.2302.12433},
  doi          = {10.48550/ARXIV.2302.12433},
  eprinttype    = {arXiv}
}

@article{DBLP:journals/corr/abs-2301-02195,
  author       = {Garett Cunningham and
                  Razvan C. Bunescu and
                  David Juedes},
  title        = {Towards Autoformalization of Mathematics and Code Correctness: {E}xperiments with Elementary Proofs},
  journal      = {CoRR},
  volume       = {abs/2301.02195},
  year         = {2023},
  url          = {https://doi.org/10.48550/arXiv.2301.02195},
  doi          = {10.48550/ARXIV.2301.02195},
  eprinttype    = {arXiv}
}

@book{Coq,
  author       = {Yves Bertot and
                  Pierre Cast{\'{e}}ran},
  title        = {Interactive Theorem Proving and Program Development - {Coq'Art}: {T}he Calculus of Inductive Constructions},
  series       = {Texts in Theoretical Computer Science.  {EATCS} Series},
  publisher    = {Springer},
  year         = {2004},
  url          = {https://doi.org/10.1007/978-3-662-07964-5},
  doi          = {10.1007/978-3-662-07964-5},
  isbn         = {978-3-642-05880-6}
}

@article{DBLP:journals/cacm/SeshiaSS22,
  author       = {Sanjit A. Seshia and
                  Dorsa Sadigh and
                  S. Shankar Sastry},
  title        = {Toward verified artificial intelligence},
  journal      = {Communications of the  {ACM}},
  volume       = {65},
  number       = {7},
  pages        = {46--55},
  year         = {2022},
  url          = {https://doi.org/10.1145/3503914},
  doi          = {10.1145/3503914},
  bibsource    = {dblp computer science bibliography, https://dblp.org}
}

@article{DBLP:journals/elektrik/KanburogluT24,
  author       = {Ali Bugra Kanburoglu and
                  Faik Boray Tek},
  title        = {Text-to-{SQL}: {A} methodical review of challenges and models},
  journal      = {Turkish Journal of Electrical Engineering and Computer Science},
  volume       = {32},
  number       = {3},
  pages        = {403--419},
  year         = {2024},
  url          = {https://doi.org/10.55730/1300-0632.4077},
  doi          = {10.55730/1300-0632.4077},
  bibsource    = {dblp computer science bibliography, https://dblp.org}
}

@inproceedings{DBLP:conf/time/BrunelloMR19,
  author       = {Andrea Brunello and
                  Angelo Montanari and
                  Mark Reynolds},
  title        = {Synthesis of {LTL} Formulas from Natural Language Texts: State of the Art and Research Directions},
  booktitle    = {26th International Symposium on Temporal Representation and Reasoning (TIME)},
  series       = {LIPIcs},
  volume       = {147},
  pages        = {17:1--17:19},
  publisher    = {Schloss Dagstuhl - Leibniz-Zentrum f{\"{u}}r Informatik},
  year         = {2019},
  url          = {https://doi.org/10.4230/LIPIcs.TIME.2019.17},
  doi          = {10.4230/LIPICS.TIME.2019.17},
  bibsource    = {dblp computer science bibliography, https://dblp.org}
}

@inproceedings{DBLP:conf/fmcad/MendozaHT24,
  author       = {Daniel Mendoza and
                  others},
  title        = {Translating Natural Language to Temporal Logics with Large Language Models and Model Checkers},
  booktitle    = { conference on Formal Methods in Computer-Aided Design (FMCAD)},
  pages        = {1--11},
  publisher    = {{IEEE}},
  year         = {2024},
  url          = {https://doi.org/10.34727/2024/isbn.978-3-85448-065-5\_17},
  doi          = {10.34727/2024/ISBN.978-3-85448-065-5\_17},
  bibsource    = {dblp computer science bibliography, https://dblp.org}
}

@article{DBLP:journals/corr/abs-2409-16461,
  author       = {Ramya Keerthy Thatikonda and
                  Jiuzhou Han and
                  Wray L. Buntine and
                  Ehsan Shareghi},
  title        = {Strategies for Improving {NL-to-FOL} Translation with {LLMs}: {D}ata Generation, Incremental Fine-Tuning, and Verification},
  journal      = {CoRR},
  volume       = {abs/2409.16461},
  year         = {2024},
  url          = {https://doi.org/10.48550/arXiv.2409.16461},
  doi          = {10.48550/ARXIV.2409.16461},
  eprinttype    = {arXiv},
  eprint       = {2409.16461},
  bibsource    = {dblp computer science bibliography, https://dblp.org}
}

@inproceedings{DBLP:conf/emnlp/Abzianidze17,
  author       = {Lasha Abzianidze},
  title        = {{LangPro}: {N}atural Language Theorem Prover},
  booktitle    = { 2017 Conference on Empirical Methods in Natural Language Processing (EMNLP)},
  pages        = {115--120},
  publisher    = {Association for Computational Linguistics},
  year         = {2017},
  url          = {https://doi.org/10.18653/v1/d17-2020},
  doi          = {10.18653/V1/D17-2020},
  bibsource    = {dblp computer science bibliography, https://dblp.org}
}

@inproceedings{DBLP:conf/mlcw/BosM05,
  author       = {Johan Bos and
                  Katja Markert},
  title        = {Recognising Textual Entailment with Robust Logical Inference},
  booktitle    = { 1st PASCAL Workshop on Machine Learning Challenges (MLCW)},
  series       = {Lecture Notes in Computer Science},
  volume       = {3944},
  pages        = {404--426},
  publisher    = {Springer},
  year         = {2005},
  url          = {https://doi.org/10.1007/11736790\_23},
  doi          = {10.1007/11736790\_23},
  bibsource    = {dblp computer science bibliography, https://dblp.org}
}

@inproceedings{DBLP:conf/uai/ZettlemoyerC05,
  author       = {Luke S. Zettlemoyer and
                  Michael Collins},
  title        = {Learning to Map Sentences to Logical Form: {S}tructured Classification with Probabilistic Categorial Grammars},
  booktitle    = { 21st Conference in Uncertainty in Artificial Intelligence (UAI)},
  pages        = {658--666},
  publisher    = {{AUAI}},
  year         = {2005},
  url          = {https://dslpitt.org/uai/displayArticleDetails.jsp?mmnu=1\&smnu=2\&article\_id=1209\&proceeding\_id=21},
  bibsource    = {dblp computer science bibliography, https://dblp.org}
}

@inproceedings{38d9186591d147e8ba3caec40080d1c4,
    author = {Oleksii Levkovskyi and Wei Li},
    title = {Generating predicate logic expressions from natural language},
    booktitle = { 2021 IEEE SoutheastCon},
    publisher = {Institute of Electrical and Electronics Engineers Inc.},
    year = {2021},
}

@inproceedings{DBLP:conf/coling/LuLGT0HXW22,
  author       = {Xuantao Lu and
                  Jingping Liu and
                  others},
  title        = {Parsing Natural Language into Propositional and First-Order Logic with Dual Reinforcement Learning},
  booktitle    = { 29th International Conference on Computational Linguistics (COLING)},
  pages        = {5419--5431},
  publisher    = {ICCL},
  year         = {2022},
  url          = {https://aclanthology.org/2022.coling-1.481},
  bibsource    = {dblp computer science bibliography, https://dblp.org}
}

@inproceedings{DBLP:conf/acl/CaoZLLY19,
  author       = {Ruisheng Cao and
                  Su Zhu and
                  Chen Liu and
                  Jieyu Li and
                  Kai Yu},
  title        = {Semantic Parsing with Dual Learning},
  booktitle    = { 57th Conference of the Association for Computational Linguistics (ACL)},
  pages        = {51--64},
  publisher    = {Association for Computational Linguistics},
  year         = {2019},
  url          = {https://doi.org/10.18653/v1/p19-1007},
  doi          = {10.18653/V1/P19-1007},
  bibsource    = {dblp computer science bibliography, https://dblp.org}
}

@inproceedings{DBLP:conf/emnlp/PanAWW23,
  author       = {Liangming Pan and
                  Alon Albalak and
                  others},
  title        = {Logic-{LM}: {E}mpowering Large Language Models with Symbolic Solvers for
                  Faithful Logical Reasoning},
  booktitle    = { 2023 Conference on Empirical Methods in Natural Language Processing (EMNLP)},
  pages        = {3806--3824},
  publisher    = {Association for Computational Linguistics},
  year         = {2023},
  url          = {https://doi.org/10.18653/v1/2023.findings-emnlp.248},
  doi          = {10.18653/V1/2023.FINDINGS-EMNLP.248},
  bibsource    = {dblp computer science bibliography, https://dblp.org}
}

@inproceedings{DBLP:conf/emnlp/OlaussonGLZSTL23,
  author       = {Theo Olausson and
                  Alex Gu and
                  Benjamin Lipkin and
                  Cedegao E. Zhang and
                  others},
  title        = {{LINC:} {A} Neurosymbolic Approach for Logical Reasoning by Combining
                  Language Models with First-Order Logic Provers},
  booktitle    = { 2023 Conference on Empirical Methods in Natural
                  Language Processing (EMNLP)},
  pages        = {5153--5176},
  publisher    = {Association for Computational Linguistics},
  year         = {2023},
  url          = {https://doi.org/10.18653/v1/2023.emnlp-main.313},
  doi          = {10.18653/V1/2023.EMNLP-MAIN.313},
  bibsource    = {dblp computer science bibliography, https://dblp.org}
}

@inproceedings{DBLP:conf/nips/YeCDD23,
  author       = {Xi Ye and
                  Qiaochu Chen and
                  Isil Dillig and
                  Greg Durrett},
  title        = {{SatLM}: {S}atisfiability-Aided Language Models Using Declarative Prompting},
  booktitle    = { 37th Annual Conference on Neural Information Processing Systems (NeurIPS)},
  year         = {2023},
  url          = {http://papers.nips.cc/paper\_files/paper/2023/hash/8e9c7d4a48bdac81a58f983a64aaf42b-Abstract-Conference.html},
  bibsource    = {dblp computer science bibliography, https://dblp.org}
}

@inproceedings{DBLP:conf/emnlp/TianLCX0J21,
  author       = {Jidong Tian and
                  Yitian Li and
                  Wenqing Chen and
                  others},
  title        = {Diagnosing the First-Order Logical Reasoning Ability Through {LogicNLI}},
  booktitle    = { 2021 Conference on Empirical Methods in Natural Language Processing (EMNLP)},
  pages        = {3738--3747},
  publisher    = {Association for Computational Linguistics},
  year         = {2021},
  url          = {https://doi.org/10.18653/v1/2021.emnlp-main.303},
  doi          = {10.18653/V1/2021.EMNLP-MAIN.303},
  bibsource    = {dblp computer science bibliography, https://dblp.org}
}

@book{Paris_Vencovská_2015, place={Cambridge}, series={Perspectives in Logic}, title={Pure Inductive Logic}, publisher={Cambridge University Press}, author={Paris, Jeffrey and Vencovská, Alena}, year={2015}, collection={Perspectives in Logic}}

@inproceedings{DBLP:conf/acl/PapineniRWZ02,
  author       = {Kishore Papineni and
                  Salim Roukos and
                  others},
  title        = {Bleu: {A} Method for Automatic Evaluation of Machine Translation},
  booktitle    = { 40th Annual Meeting of the Association for Computational Linguistics (ACL)},
  pages        = {311--318},
  publisher    = {{ACL}},
  year         = {2002},
  url          = {https://aclanthology.org/P02-1040/},
  doi          = {10.3115/1073083.1073135},
  bibsource    = {dblp computer science bibliography, https://dblp.org}
}

@incollection{DBLP:reference/mc/BarrettT18,
  author       = {Clark W. Barrett and
                  Cesare Tinelli},
  title        = {Satisfiability Modulo Theories},
  booktitle    = {Handbook of Model Checking},
  pages        = {305--343},
  publisher    = {Springer},
  year         = {2018},
  url          = {https://doi.org/10.1007/978-3-319-10575-8\_11},
  doi          = {10.1007/978-3-319-10575-8\_11},
  bibsource    = {dblp computer science bibliography, https://dblp.org}
}

@book{DBLP:series/faia/336,
  editor       = {Armin Biere and
                  Marijn Heule and
                  Hans van Maaren and
                  Toby Walsh},
  title        = {Handbook of Satisfiability - Second Edition},
  series       = {Frontiers in Artificial Intelligence and Applications},
  volume       = {336},
  publisher    = {{IOS} Press},
  year         = {2021},
  url          = {https://doi.org/10.3233/FAIA336},
  doi          = {10.3233/FAIA336},
  isbn         = {978-1-64368-160-3},
  bibsource    = {dblp computer science bibliography, https://dblp.org}
}

@article{du2024short,
  author       = {Rick Du and
                  Huilong An and
                  Keyu Wang and
                  Weidong Liu},
  title        = {A Short Review for Ontology Learning from Text: Stride from Shallow
                  Learning, Deep Learning to Large Language Models Trend},
  journal      = {CoRR},
  volume       = {abs/2404.14991},
  year         = {2024},
  url          = {https://doi.org/10.48550/arXiv.2404.14991},
  doi          = {10.48550/ARXIV.2404.14991},
  eprinttype    = {arXiv},
  eprint       = {2404.14991},
  bibsource    = {dblp computer science bibliography, https://dblp.org}
}

@article{ARMARY2025100693,
title = {Ontology learning towards expressiveness: A survey},
journal = {Computer Science Review},
volume = {56},
pages = {100693},
year = {2025},
issn = {1574-0137},
doi = {https://doi.org/10.1016/j.cosrev.2024.100693},
url = {https://www.sciencedirect.com/science/article/pii/S1574013724000765},
author = {Pauline Armary and Cheikh Brahim El-Vaigh and Ouassila {Labbani Narsis} and Christophe Nicolle}
}

@inproceedings{Dellschaft2006,
  author       = {Klaas Dellschaft and
                  Steffen Staab},
  title        = {On How to Perform a Gold Standard Based Evaluation of Ontology Learning},
  booktitle    = {5th International Semantic Web Conference,
                  {(ISWC)}},
  series       = {Lecture Notes in Computer Science},
  volume       = {4273},
  pages        = {228--241},
  publisher    = {Springer},
  year         = {2006},
  url          = {https://doi.org/10.1007/11926078\_17},
  doi          = {10.1007/11926078\_17},
  bibsource    = {dblp computer science bibliography, https://dblp.org}
}

@article{DBLP:journals/corr/abs-2505-09388,
  author       = {An Yang
                  and others},
  title        = {Qwen3 Technical Report},
  journal      = {CoRR},
  volume       = {abs/2505.09388},
  year         = {2025},
  url          = {https://doi.org/10.48550/arXiv.2505.09388},
  doi          = {10.48550/ARXIV.2505.09388},
  eprinttype    = {arXiv},
  eprint       = {2505.09388},
  bibsource    = {dblp computer science bibliography, https://dblp.org}
}
